\documentclass[10pt,a4paper]{article}

\usepackage{arxiv}
\usepackage[utf8]{inputenc}
\usepackage[T1]{fontenc}
\usepackage{hyperref}       
\usepackage{url}            
\usepackage{booktabs}       
\usepackage{nicefrac}       
\usepackage{microtype}      
\usepackage{graphicx}
\usepackage{geometry}
\usepackage{algorithmic}
\usepackage{textcomp}
\usepackage{comment}
\def\BibTeX{{\rm B\kern-.05em{\sc i\kern-.025em b}\kern-.08em
T\kern-.1667em\lower.7ex\hbox{E}\kern-.125emX}}

\usepackage{amsmath,amssymb,amsfonts}
\usepackage[dvipsnames]{xcolor}   
\usepackage[braket, qm]{qcircuit} 
\usepackage{subcaption}           
\usepackage{float}                
\usepackage{tikz}                 
\usepackage{colortbl}             
\usepackage{savesym}              
\usepackage{physics}              
\usepackage[htt]{hyphenat}        
\usepackage{soul}                 
\usepackage{enumitem}             
\usepackage{etoolbox}             
\usepackage{makecell}             
\usepackage{array}                
\usepackage[english]{babel}
\usetikzlibrary{fit}
\usetikzlibrary{positioning}

\renewcommand\ket[1]{\left|#1\right\rangle}
\renewcommand\bra[1]{\left\langle#1\right|}
\renewcommand\vec[1]{\boldsymbol{\mathbf{#1}}} 

\definecolor{zhawblue}{rgb}{0.00, 0.39, 0.65}
\definecolor{codegreen}{rgb}{0,0.6,0}
\definecolor{codegray}{rgb}{0.5,0.5,0.5}
\definecolor{codepurple}{rgb}{0.58,0,0.82}
\definecolor{codebackground}{rgb}{0.93,0.94,0.95}
\definecolor{darkgreen}{rgb}{0.0, 0.4, 0.0}
\definecolor{softyellow}{rgb}{1.0, 1.0, 0.4}
\definecolor{softblue}{rgb}{0.6, 0.6, 1.0}
\definecolor{softgreen}{rgb}{0.4, 1.0, 0.4}
\definecolor{MyTeal}{rgb}{0, 0.792, 0.388}
\definecolor{classical}{RGB}{159, 6, 17}
\definecolor{noisy}{RGB}{0, 89, 137}
\definecolor{ideal}{RGB}{0, 97, 69}

\makeatletter
\pretocmd{\tagform@}{\color{gray}}{}{}
\makeatother

\usepackage[colorinlistoftodos,prependcaption,textsize=tiny]{todonotes}

\title{Quanvolutional Neural Networks for Spectrum Peak-Finding}

\author{Lukas Bischof \\
Institute of Computer Science \\
Zurich University of Applied Sciences \\
\texttt{me@luk4s.dev} \\
\And
Rudolf M. Füchslin \\
Institute of Applied Mathematics and Physics \\
Zurich University of Applied Sciences \\
European Centre for Living Technology \\
University Ca'Foscari, Venice, Italy \\
\texttt{furu@zhaw.ch} \\
\And
Kurt Stockinger \\
Institute of Computer Science \\
Zurich University of Applied Sciences \\
\texttt{stog@zhaw.ch} \\
\And
Pavel Sulimov* \\
Institute of Computer Science \\
Zurich University of Applied Sciences \\
\texttt{suli@zhaw.ch} \\
}

\renewcommand{\headeright}{}
\renewcommand{\undertitle}{}

\hypersetup{
    pdftitle={Quanvolutional Neural Networks for Spectrum Peak Finding},
    pdfauthor={Lukas Bischof},
    pdfkeywords={Quantum Computing, NMR, Peak Finding, Deconvolution, Experimental Evaluation},
}

\begin{document}
    \pagenumbering{gobble}
    \pagenumbering{arabic}
    \twocolumn[%
        \begin{@twocolumnfalse}
            \maketitle
            \begin{abstract}
                The analysis of spectra, such as Nuclear Magnetic Resonance (NMR) spectra, for the comprehensive characterization of peaks
                is a challenging task for both experts and machines, especially with complex molecules.
                This process, also known as deconvolution, involves identifying and quantifying the peaks in the spectrum.
                Machine learning techniques have shown promising results in automating this process.
                With the advent of quantum computing, there is potential to further enhance these techniques.
                In this work, inspired by the success of classical Convolutional Neural Networks (CNNs),
                we explore the use of Quanvolutional Neural Networks (QuanvNNs) for the multi-task peak finding problem,
                involving both peak counting and position estimation.
                We implement a simple and interpretable QuanvNN architecture that can be directly compared to its classical CNN counterpart,
                and evaluate its performance on a synthetic NMR-inspired dataset.
                Our results demonstrate that QuanvNNs outperform classical CNNs on challenging spectra,
                achieving an 11\% improvement in F1 score and a 30\% reduction in mean absolute error for peak position estimation.
                Additionally, QuanvNNs appear to exhibit better convergence stability for harder problems.
            \end{abstract}
            \vspace{2em}
        \end{@twocolumnfalse}%
    ]

    \section{Introduction}\label{sec:introduction}
\begin{figure}[h]
    \centering
    \includegraphics[width=0.85\columnwidth]{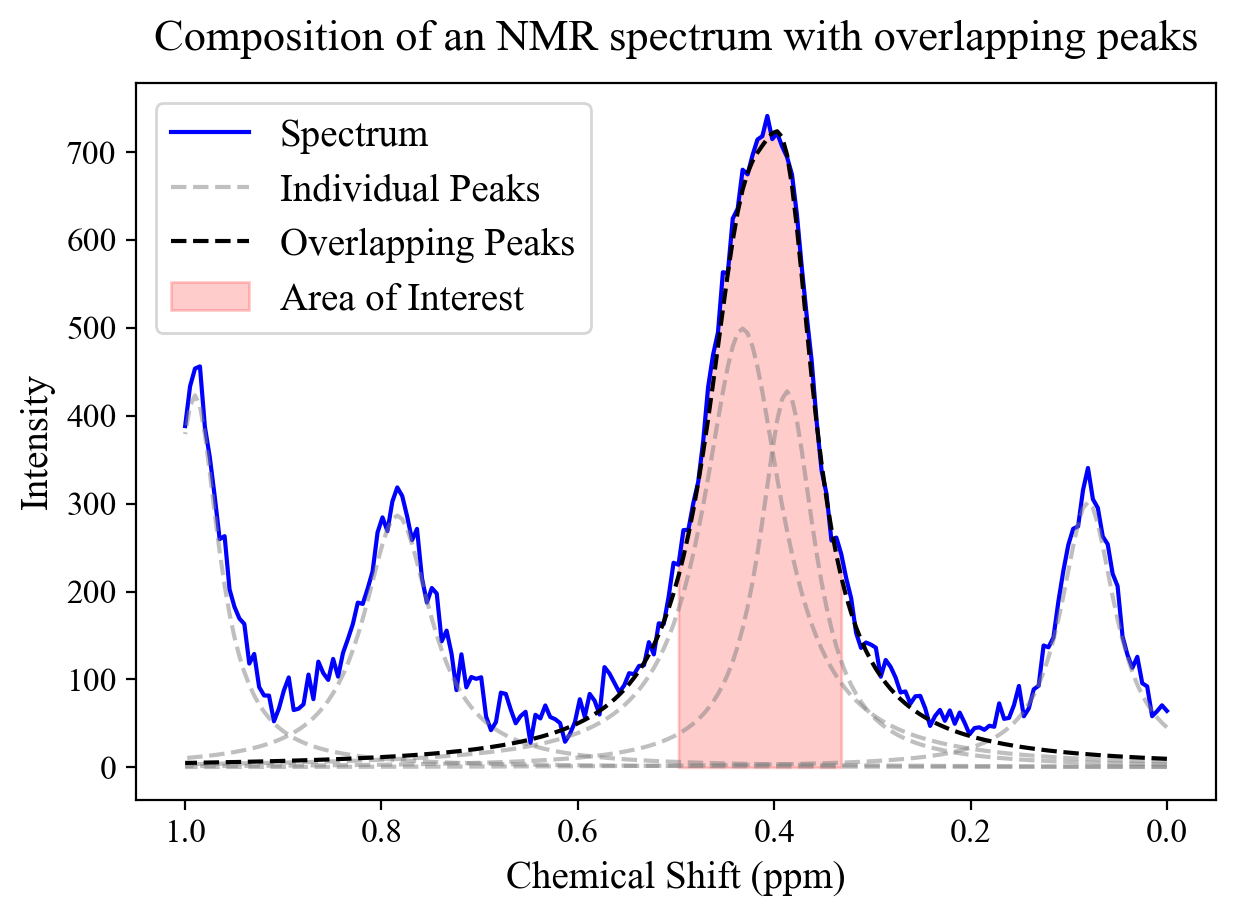}
    \caption{
        Deconvolution of a spectrum is an ill-posed inverse problem.
        In this example, the original spectrum (blue) is constructed from individual peaks (gray), forming the ground truth.
        However, if two peaks are too close together, they could be reconstructed as a single broader peak (black) instead.
        Similarly, a single peak could also be reconstructed as two peaks, leading to an overfitting of the spectrum.
    }
    \label{fig:ill-posedness}
\end{figure}

    Interpretation of spectra represents a fundamental challenge in analytical sciences, particularly for complex molecular systems (like Nuclear Magnetic Resonance spectroscopy~\cite{peak-picking-nmr-spectral-data-2014} (NMR)
or Mass Spectrometry~\cite{evaluation-of-peak-picking-algorithms-2010} (MS)) where traditional peak picking algorithms face limitations. While classical methods like Scipy Peak Finder~\cite{scipy-2020} suffice for simple spectra, they struggle with overlapping peaks, low signal-to-noise ratios, and the ill-posed inverse problem inherent to spectral deconvolution (Fig.~\ref{fig:ill-posedness}). The emergence of quantum machine learning (QML)~\cite{quantum-machine-learning-2017, simoes2023qml, frehner2024autoenc} offers new avenues to address these challenges by leveraging the high-dimensional Hilbert space and entanglement properties of quantum systems.

Mathematically, the peak finding problem can be formulated as an optimization task in a high-dimensional space where quantum systems naturally operate. The quantum feature maps implemented through quanvolutional layers project classical data into exponentially larger Hilbert spaces, potentially enabling better separation of overlapping spectral features that challenge classical methods.

We focus on NMR spectroscopy to introduce the peak finding problem.
NMR is a powerful analytical technique used to elucidate molecular structure.
An NMR spectrum consists of peaks, each representing a frequency at which atomic nuclei resonate in a magnetic field. These resonance frequencies and in consequence the position and detailed shape of the peaks in the spectrum are given by the properties of the resonating atoms but are varied by the chemical surrounding of these atoms.
The number of peaks can range from a few to hundreds, depending on molecular complexity.

Gaining insights from the spectrum requires identifying and quantifying these peaks through a process known as deconvolution,
which is fundamental for facilitating workflows in various fields,
such as chemistry and life sciences~\cite{deep-learning-in-nmr-2020}.
During deconvolution, each peak is characterized by its center position
(frequency coordinates according to chemical shift), shape, amplitude, and other
properties such as coupling constants.
Robust peak finding is among the biggest automation challenges and forms the foundation for subsequent analysis.
Difficulties arise from overlapping peaks with varying shapes, low signal-to-noise ratios (SNR),
spectral distortions, and artifacts.
Moreover, peak finding is an ill-posed inverse problem~\cite{inverse-methods-in-nmr-2003},
meaning that multiple valid solutions can exist for a given spectrum, as illustrated in~\autoref{fig:ill-posedness}.
Preferring sparse solutions with fewer peaks may fail to accurately reconstruct the original
spectrum, as demonstrated in~\autoref{fig:ill-posedness}, where the sparse solution would yield a four-peak
reconstruction instead of the correct five peaks.
Conversely, if one refrains from favoring sparse solutions, one may be confronted with overfitting, where more peaks are detected than actually present, despite achieving low reconstruction error
relative to the original spectrum.

Convolutional neural networks (CNNs) have shown promising results for
automating peak finding~\cite{deep-picker-2021, deconvolution-of-1d-nmr-spectra-2023}.
Convolutions can be thought of as a ``stack'' of transformations on the input data,
where each layer iteratively extracts different spatially-local features from the input data.
The resulting output is a tensor of $n$ feature maps that provide abstract
representations of learned local features~\cite{gradient-based-learning-2002}.
Since the learned filters use the same weights for each local region, the transformation is translation invariant,
enabling detection of the same features regardless of their spatial position in the input.

In quantum machine learning, an equivalent to CNNs exists,
known as quantum convolutional neural networks (QNN)~\cite{quantum-convolutional-neural-networks-2019}.
These networks replicate classical CNN structures entirely using quantum circuits.
However, the required qubit count scales with input dimensionality,
potentially exceeding current quantum hardware capabilities for higher-resolution spectra.

A promising alternative for the noisy intermediate-scale quantum (NISQ) era is the
quanvolutional neural network (QuanvNN) approach.
QuanvNNs were first introduced by~
\cite{quanvolutional-neural-networks-2020}, who employed random quantum circuits to process input data
through local transformations.
This hybrid quantum-classical approach (e.g. \cite{bischof2025entity}) incorporates a quantum input layer followed by
classical layers in a unified model.
Similar to classical convolutions, the quantum convolution filter operates as a sliding window, applying
transformations to local input regions using quantum circuits to produce multiple output channels.
This design allows for more efficient quantum resource utilization compared to QNNs, as qubit count scales with
kernel size rather than input dimensionality.

The structural similarity between QuanvNNs and classical CNNs enables direct performance and interpretability comparisons, making them suitable candidates for investigating whether quantum transformations
can provide advantages over classical methods in this domain.
A detailed description of the architecture and the quantum circuits used in this work is provided in
Section~\ref{subsec:approach-quanvolutional-neural-networks}.

Motivated by the success of CNNs in this problem domain, we investigate the feasibility of applying
\emph{quanvolutional neural networks} to the peak finding problem.
The contributions of our paper are as follows:

\begin{enumerate}
    \item{
        We construct a synthetic dataset inspired by NMR spectra,
        allowing us to have better control over the characteristics of the peaks.
        The dataset considers the scaling limitations of current noisy intermediate-scale quantum hardware.
        The generated spectra are therefore small enough to be processed on current quantum hardware,
        while still retaining the challenging characteristics such as overlapping peaks, low SNR, and distortions.
    }
    \item{
        We implement a simple and interpretable neural network architecture using a pluggable input layer,
        which enables us to evaluate classical convolutional and quantum convolutional layers.
        We can therefore directly compare the performance of the QuanvNN and the classical CNN.
    }
    \item{
        We evaluate the performance of multiple QuanvNNs using different quanvolution kernels on the synthetic dataset and compare it to a classical CNN.
        The results suggest that the QuanvNN has better convergence properties and lower error rates than the classical CNN for harder problems.
    }
\end{enumerate}

To the best of our knowledge, this is the first work that explores the use of QuanvNNs for the peak finding problem in NMR spectroscopy.

    \section{Fundamentals of Quantum Computing and Quanvolutional Neural Networks}
    \label{sec:fundamentals-of-quantum-computing}
    In this section, we introduce the fundamental concepts of quantum computing.
Furthermore, for an in-depth introduction to quantum computing,
we refer the reader to the literature~\cite{quantum-computation-and-quantum-information-2001}
as we will only cover the essential concepts most relevant to this work.

\subsection{Quantum Computing}\label{subsec:quantum-computing}
The qubit is the fundamental unit of quantum information, analogous to the classical bit.
Unlike a classical bit, a qubit is associated with a two-state quantum system (e.g., a spin-1/2 particle)
that can exist in superposition and is described by a complex vector in a two-dimensional Hilbert space.
Using Dirac notation, we can represent the two base states of a qubit as $\ket{0}$ and $\ket{1}$, forming an
orthonormal basis.
A general state of a qubit $\ket{\psi}$ can be expressed as a linear combination of these two states:
$\ket{\psi} = \alpha \ket{0} + \beta \ket{1}$,
where $\alpha$ and $\beta$ are complex numbers satisfying the normalization condition $|\alpha|^2 + |\beta|^2 = 1$.
To manipulate qubits, we use so-called quantum gates, which are unitary operations that transform the state of a qubit.
Finally, to retrieve information from a qubit, we perform a measurement which collapses the qubit's state to one
of the basis states $\ket{0}$ or $\ket{1}$ with probabilities $|\alpha|^2$ and $|\beta|^2$, respectively.
Importantly, we cannot directly observe the coefficients $\alpha$ and $\beta$, but can only infer them
from the probabilities obtained upon measurement.

Qubits can be combined to form quantum registers, which live in a joint Hilbert space formed by the tensor product
of the individual qubit spaces:
$\mathcal{H} = \mathcal{H}_1 \otimes \mathcal{H}_2 \otimes \ldots \otimes \mathcal{H}_n$,
where $\mathcal{H}_i$ is the Hilbert space of the $i$-th qubit.
For instance, two independent qubits $\ket{\psi_1}$ and $\ket{\psi_2}$ can be combined to form a quantum register
$\ket{\psi} = \ket{\psi_1} \otimes \ket{\psi_2}$.

The quantum system can be manipulated using quantum gates, which are unitary operations acting on the full quantum
register.
These can be represented as a matrix $U$ fulfilling the condition $U^\dagger U = I = U U^\dagger$,
where $U^\dagger$ is the conjugate transpose of $U$ and $I$ is the identity matrix.
As a result, the state of the quantum register is transformed as follows:

\begin{equation}
    \ket{\psi'} = U \ket{\psi}
    \label{eq:quantum-gate}
\end{equation}

We refer to this as gate-based quantum computing when applying one or more unitary operations (i.e., quantum gates)
to a quantum register.
Quantum gates can furthermore be parameterized by a set of parameters $\theta$, allowing us to
control the transformation of the quantum state.
These so-called variational quantum circuits are particularly useful for machine learning algorithms, where we
fine-tune the parameters to minimize a cost function.

\subsection{Quanvolutional Neural Networks}\label{subsec:quanvolutional-neural-networks-fundamentals}

\begin{figure}[h]
    \centering
    \includegraphics[width=\columnwidth]{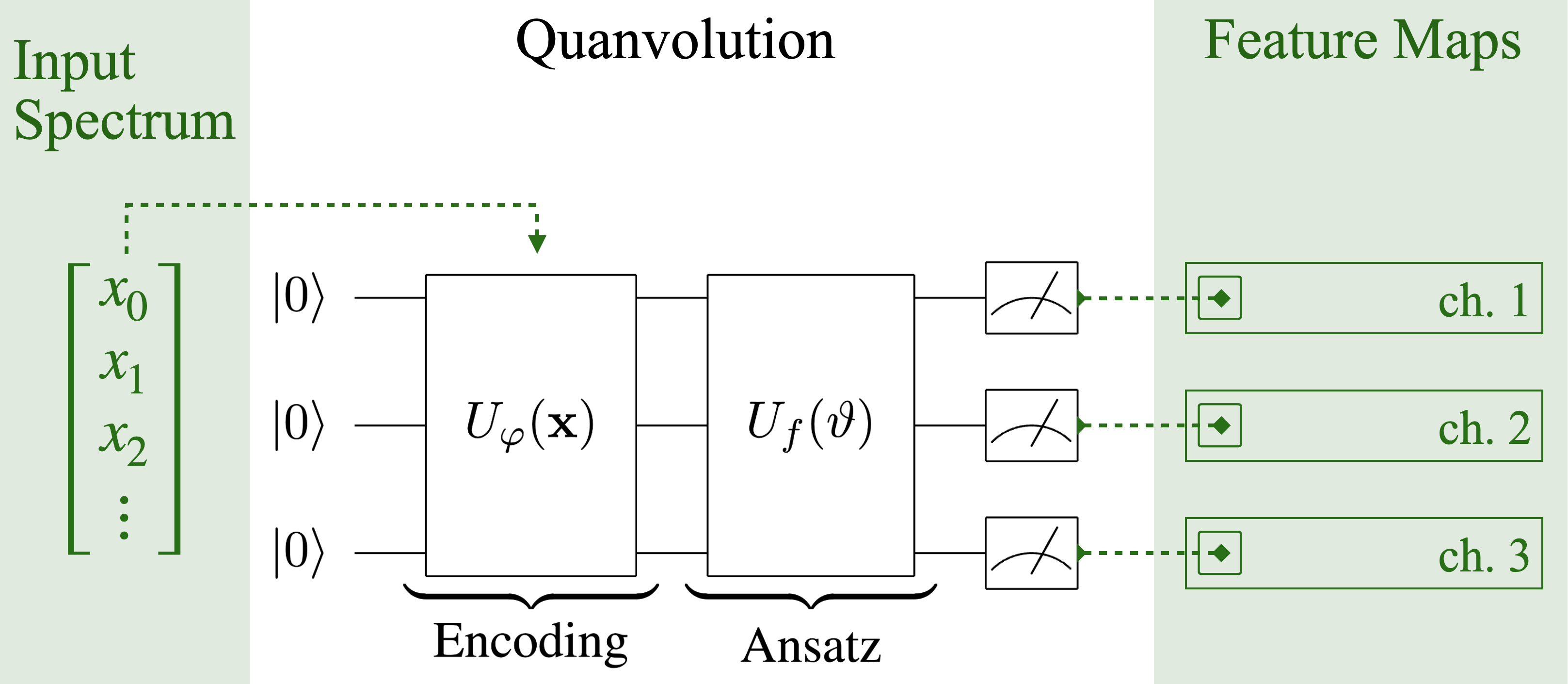}
    \caption{
        Schematic representation of a quantum convolution (quanvolution) using a three qubit kernel.
        Green represents classical data.
    }
    \label{fig:quanvolution-scheme}
\end{figure}

Our approach to Quanvolutional Neural Networks (QuanvNNs) can be directly derived from classical CNNs.
A schematic representation is shown in~\autoref{fig:quanvolution-scheme}, where we illustrate a generalized
three-qubit kernel for simplicity.

We first examine the quantum circuit that implements the quantum convolution operation.
The region of the input data to which the kernel is applied is constrained by the number of qubits used.
Given that we have $n_q$ qubits in the kernel,
the number of data points that can be processed is at most $2^{n_q}$.
The operator $U_\varphi(\vec{x})$ then transforms the classical data $\vec{x}$ into a quantum state
$\ket{\varphi(\vec{x})} = U_\varphi(\vec{x}) \ket{0}^{\otimes n_q}$ of
Hilbert space $\mathbb{H}^{n_q}$.
The ansatz $U_f(\vartheta)$ with parameters $\vartheta$, which can be tuned by a classical optimizer, is then applied to
the quantum state $\ket{\psi}$.
Measurement of the expectation values of the final quantum state with respect to the observable $\hat{O}$ yields a classical
output vector $\vec{y}$, which can be interpreted as the output of the quantum convolution operation:

\begin{equation}
    y = f(\vec{x}, \vec{\vartheta}) =
    \bra{\varphi(\vec{x})} U^\dagger_f(\vec{\vartheta}) \hat{O} U_f(\vec{\vartheta}) \ket{\varphi(\vec{x})}
    \label{eq:quanvolution-measurement}
\end{equation}

To optimize the parameters $\vartheta$ of the ansatz, we need to compute the gradient with respect to these parameters. If we specify the ansatz further by assuming
\begin{equation}
   U_f(\vec{\vartheta}) = U_{N-1}(\vartheta_{N-1}) ... U_{0}(\vartheta_{0})
    \label{eq:quanvolution-gradient-ansatz1}
\end{equation}
with ($P_i$ one of the Pauli-gates)
\begin{equation}
   U_j(\vartheta_j) = \exp(-i \frac{\vartheta_j}{2} P_j),
    \label{eq:quanvolution-gradient-ansatz2}
\end{equation}
we can apply the parameter-shift rule for estimating the gradient (\cite{mitarai2018quantum}). For each component $\vartheta_j$ of the parameter vector $\vec{\vartheta}$, the gradient is given by:

\begin{equation}
    \frac{\partial f(\vec{x}, \vec{\vartheta})}{\partial \vartheta_j}
    = \frac{1}{2} \left( f(\vec{x}, \vec{\vartheta} + \frac{\pi}{2}\vec{e}_j) - f(\vec{x}, \vec{\vartheta} - \frac{\pi}{2}\vec{e}_j) \right)
    \label{eq:quanvolution-gradient}
\end{equation}

where $\vec{e}_j$ is the $j$-th unit vector. This parameter-shift rule holds specifically for unitary operators of the form $U_j(\vartheta_j) = \exp(-i \frac{\vartheta_j}{2} P_j)$ with Pauli operators $P_j$, as used in our ansätze (see Section~\ref{subsec:approach-quanvolutional-neural-networks}). In our implementation using PennyLane, this relation is satisfied for the ansätze employed in this work.


Other techniques are also available,
such as adjoint differentiation~\cite{efficient-calculation-of-gradients-2020},
which exploits the reversibility of quantum circuits when using a simulator, or finite differences.
We will not elaborate on these techniques here, as our requirement is simply that a gradient algorithm
exists, regardless of which specific method is used.

The mathematical formulation of the quantum convolution operation can be derived from the classical convolution.
Let us assume that we have a classical input signal $\vec{x} = (x_0, x_1, \ldots, x_{n-1})$ with $\vec{x} \in \mathbb{R}^n$,
and a kernel $\vec{w} = (w_0, w_1, \ldots, w_{k - 1})$ with $\vec{w} \in \mathbb{R}^k$, where typically $k \ll n$.
We further assume a convolution with a stride of one, no padding, and dilation of one, as this
is what we will use in our QuanvNNs implementation.
The classical one-dimensional convolution operation is defined as:

\begin{equation}
    y_i = \sum_{j=0}^{k-1} w_j \cdot x_{i + j} + b
    \label{eq:classical-convolution}
\end{equation}

where $i = 0, 1, \ldots, n - k$ and $b$ is a bias term.
For the quanvolutional operation, we use the same formulation, but replace the classical kernel $\vec{w}$
with a quantum kernel $f(\vec{x}, \vec{\vartheta})$, as introduced in~\autoref{eq:quanvolution-measurement}:

\begin{align}
    y_i & = f(\vec{x}_{i:i+k-1}, \vec{\vartheta}) + b \\
    & = \bra{\varphi(\vec{x}_{i:i+k-1})} U^\dagger_f(\vec{\vartheta}) \hat{O} U_f(\vec{\vartheta}) \ket{\varphi(
    \vec{x}_{i:i+k-1})} + b
    \label{eq:quanvolutional-convolution}
\end{align}

where $i = 0, 1, \ldots, n - k$.
To optimize the parameters $\vec{\vartheta}$ for a given scalar loss function $\mathcal{L}$, we need to
compute the gradient of the loss function with respect to these parameters.
In the classical case, this is done by differentiating the loss function with respect to the kernel weights.
Specifically, for each weight $w_j$ in the kernel, we compute the gradient of the loss function with respect to that
weight:

\begin{equation}
    \frac{\partial \mathcal{L}}{\partial w_j} = \sum_{i=0}^{n-k} \frac{\partial \mathcal{L}}{\partial y_i} \cdot \frac{\partial y_i}{\partial w_j}
    = \sum_{i=0}^{n-k} \frac{\partial \mathcal{L}}{\partial y_i} \cdot x_{i + j}
    \label{eq:classical-convolution-gradient}
\end{equation}

Additionally, we can also compute the gradient with respect to the bias term $b$:

\begin{equation}
    \frac{\partial \mathcal{L}}{\partial b} = \sum_{i=0}^{n-k} \frac{\partial \mathcal{L}}{\partial y_i}
    \label{eq:classical-convolution-bias-gradient}
\end{equation}

In the quantum case, the gradient of the loss function with respect to the parameters $\vec{\vartheta}$ follows the chain rule:

\begin{align}
    \frac{\partial \mathcal{L}}{\partial \vartheta_k} & = \sum_{i=0}^{n-k} \frac{\partial \mathcal{L}}{\partial y_i} \cdot \frac{\partial y_i}{\partial \vartheta_k} \\
    & = \sum_{i=0}^{n-k} \frac{\partial \mathcal{L}}{\partial y_i} \cdot
    \frac{\partial f(\vec{x}_{i:i+k-1}, \vec{\vartheta})}{\partial \vartheta_k}
    \label{eq:quanvolutional-convolution-gradient}
\end{align}

The crucial component is the computation of $\partial f(\vec{x}, \vec{\vartheta})/\partial \vartheta_k$, which is evaluated using the parameter-shift rule (Equation~\ref{eq:quanvolution-gradient}) for each parameter $\vartheta_k$ in the ansatz. This requires two quantum circuit evaluations per parameter: one with $\vartheta_k + \pi/2$ and one with $\vartheta_k - \pi/2$, while keeping all other parameters fixed.

The gradient calculation for the bias term $b$ remains the same as in the classical case, as it is independent of
the quantum kernel, and is shown in~\autoref{eq:classical-convolution-bias-gradient}.

The described convolution operations all act on a single channel of the input data and output another single channel.
To increase the number of output channels, we would classically use multiple kernels $\vec{w}_k$, each producing a
separate output channel.
In the quantum case, we can use multiple observables $\hat{O}_k$ and measure the expectation values for different
qubits.
For the three-qubit example kernel shown in~\autoref{fig:quanvolution-scheme}, we could for instance use
the observables $\hat{O}_1 = \sigma_z \otimes I \otimes I$, $\hat{O}_2 = I \otimes \sigma_z \otimes I$,
and $\hat{O}_3 = I \otimes I \otimes \sigma_z$ to measure the expectation values of the three qubits using the
Pauli-Z operator $\sigma_z$, thus obtaining three output channels.

\subsection{Theoretical Foundations for Quantum Scalability}

The theoretical advantages of quantum systems become increasingly significant as problem complexity grows, due to fundamental mathematical properties, namely feature space dimensionality, entanglement and information theoretic advantages. 

\subsubsection{Feature Space Dimensionality}
Classical CNNs operate in polynomial feature spaces: for a kernel processing $n$ data points (e.g., $n$ pixels in an image or $n$ spectral measurement points) and $k$ convolutional filters, the effective feature space dimension scales as $\mathcal{O}(n \cdot k)$. Quantum circuits, by contrast, embed data into an exponentially large Hilbert spaces:

\begin{equation}
    \dim(\mathcal{H}_{\text{quantum}}) = 2^{n_q} \quad \text{vs.} \quad \dim(\mathcal{H}_{\text{classical}}) = n \cdot k
\end{equation}

where $n_q$ denotes the number of qubits. This exponential dimension provides \textit{potential} expressive power when the feature map and data structure are aligned (e.g., entanglement-induced correlations), but high dimensionality alone does not guarantee better performance and must be balanced against trainability (see, e.g., \cite{qml-hilbert-spaces-2019}).

\subsubsection{Entanglement and Correlation Capture}
The ability to capture complex correlations scales fundamentally differently between classical and quantum systems~\cite{classical-correlation-scaling-2006}. For $m$ overlapping peaks, a comparable classical block (e.g., a dense layer of width proportional to $m$) typically needs $\mathcal{O}(m^2)$ parameters to represent pairwise correlations. In our entangling quantum block, the number of tunable rotation angles scales linearly with the number of qubits, i.e., $\mathcal{O}(n_q)$ (adjust this scaling if your specific ansatz differs).

We use $n_q$ qubits to encode $n_q$ discrete spectral bins set by the chosen resolution—this is independent of the (unknown) number of peaks $m$. The encoded state is

\[
    \ket{\psi} = \sum_{i_1,\ldots,i_{n_q}} c_{i_1\ldots i_{n_q}} \ket{i_1}\otimes\cdots\otimes\ket{i_{n_q}},
\]
where the coefficients $c_{i_1\ldots i_{n_q}}$ implicitly encode correlations across bins. Peaks are represented through their sampled amplitudes over these bins and can exceed $n_q$ in count. Entangling layers then enable interference patterns that capture multi-bin correlations, potentially reducing the number of explicit parameters needed relative to classical pairwise modeling.


\subsubsection{Information-Theoretic Advantages}
From an information-theoretic perspective, quantum systems may offer advantages in certain computational tasks~\cite{quantum-advantage-information-2022}. The quantum Fisher information matrix for the strongly entangling ansatz (introduced in Section~\ref{subsec:approach-quanvolutional-neural-networks}) shows a richer structure:

\begin{equation}
    \mathcal{F}_{Q}^{ij} = 4\left[\langle\partial_i\psi|\partial_j\psi\rangle - \langle\partial_i\psi|\psi\rangle\langle\psi|\partial_j\psi\rangle\right]
    \label{eq:fisher-information}
\end{equation}

where $\partial_i = \partial/\partial\vartheta_i$ denotes the partial derivative with respect to the $i$-th ansatz parameter $\vartheta_i$. The factor of 4 arises from the normalization of the quantum Fisher information metric. The off-diagonal terms ($i \neq j$) capture correlations between parameter sensitivities, while the diagonal terms measure the information content of individual parameters. A richer structure in $\mathcal{F}_{Q}^{ij}$ indicates that the quantum circuit exhibits greater sensitivity to variations in the ansatz parameters $\vartheta_i$, which can lead to more informative gradients during optimization. However, whether this translates to better optimization landscapes depends on the specific problem and requires empirical validation, as optimization landscapes can be affected by barren plateaus~\cite{barren-plateaus-2018,cost-function-dependent-barren-plateaus-2021}.

    \section{Approach: Quanvolutional Neural Networks for NMR Peak Finding}\label{sec:approach}
    In this work, we compare two neural network versions: a quantum convolutional neural network (QuanvNN) and a classical CNN\@.
Both use identical training methods and structures, differing only in their convolutional layer implementation.
The models are designed to solve a multi-task regression problem that involves predicting both the number of peaks and their positions in a spectrum.
The spectra in our dataset contain between zero and five peaks, with positions normalized to the range $[0, 1]$. We fix the output to five peak slots to match this dataset bound; this cap is independent of the quantum kernel size.
For the QuanvNN, we implement and compare multiple quantum circuits against a classical convolutional layer.

\subsection{Quanvolutional Neural Networks}\label{subsec:approach-quanvolutional-neural-networks}

\begin{figure*}[h]
    \centering
    \includegraphics[width=\textwidth]{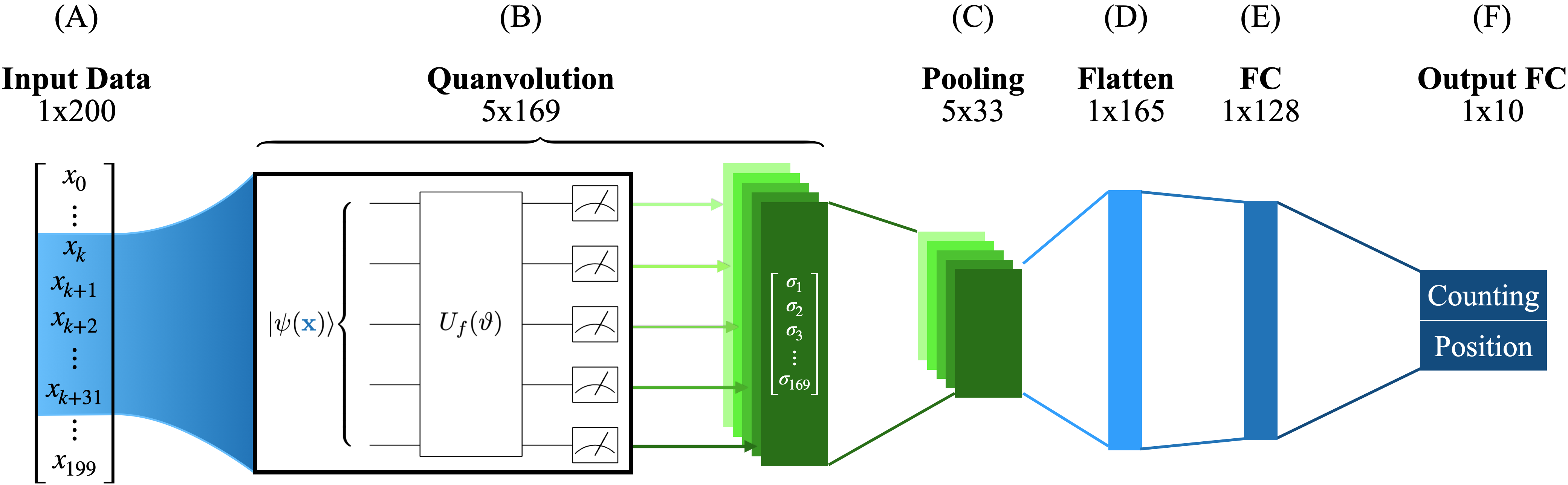}
    \caption[Architecture of the QuanvNN]{%
        Architecture of the quanvolutional neural network (QuanvNN) used in this work.
        Each spectrum consisting of 200 measurement points (A) is processed by the quantum convolution layer (B),
        which transforms the input data into five channel outputs.
        With each transformation step, the quanvolution filter takes a 32-dimensional local region of the spectrum as input.
        Since a stride size of one is used, the output per channel of the quanvolution layer is a $200-32+1=169$ dimensional tensor.
        The output of the quantum convolution layer is then processed by a max pooling layer with a kernel size of 5 (C),
        and later flattened (D) back to a one-dimensional tensor.
        Next, the flattened output is processed by a fully connected layer (E) with 128 neurons and a ReLU activation function.
        Finally, the output is produced by a fully connected layer (F) with a softmax activation function.
    }
    \label{fig:quanv-architecture}
\end{figure*}

The schematic architecture of our model implemented is illustrated in~\autoref{fig:quanv-architecture}. We will now delve into the intricacies of the architecture, detailing data flow from input to output:

\begin{enumerate}[label=(\textbf{\Alph*}), leftmargin=*]
    \item{
        \textbf{Input Data:} The model accepts a spectrum consisting of 200 measurement points, represented as a one-dimensional
        array of size 200. Each point corresponds to a \emph{normalized intensity value} ranging between 0 and 1.
    }
    \item{
        \textbf{Quanvolution:} The input is processed by a quantum convolution layer with $n_q = 5$ qubits (state space dimension $2^{n_q} = 32$), yielding a kernel that consumes 32 adjacent spectral points. The choice $n_q=5$ is a hardware/simulation and receptive-field design decision, not tied to the maximum number of peaks. The circuit implements a parameterized ansatz $U_f(\vec{\vartheta})$ that acts on an
        initial quantum state $\ket{\psi(\vec{x})}$ prepared by embedding the classical data $\vec{x}$
        via the unitary operation $U_\varphi(\vec{x})$ (see Eq.~\ref{eq:quanvolution-measurement}).
        The ansätze used in this work will be introduced later in this section.

        The initial quantum state is prepared by embedding classical data into the quantum amplitudes.
        For our 5-qubit circuits, the initial state can be expressed as:

        \begin{equation*}
            \ket{\psi(\vec{x})} = \sum_{j=0}^{2^5 - 1} \frac{1}{|\vec{x}|} x_j \ket{j},
        \end{equation*}

        where $\vec{x} = (x_0, x_1, \ldots, x_{31})$ is a 32-dimensional vector representing the spectral window currently being processed
        (i.e., 32 consecutive intensity values from the input spectrum),
        and $|\vec{x}| = \sqrt{\sum_{j=0}^{31} x_j^2}$ denotes its $\ell_2$ norm.
        The normalization ensures that the quantum state is properly normalized: $\bra{\psi(\vec{x})|\psi(\vec{x})} = 1$.
        This state preparation can be implemented through uniformly controlled rotations
        as described by~\cite{transformation-of-quantum-states-2004}.
        When using quantum simulators, arbitrary initial states can be directly set, bypassing the embedding procedure.

        After each transformation step, we measure the Pauli-Z expectation values of the final quantum state per qubit.
        Formally, for each qubit $i \in \{0, 1, 2, 3, 4\}$, we compute:
        \begin{equation*}
            \langle Z_i \rangle = \bra{\psi_{\text{out}}} Z_i \ket{\psi_{\text{out}}},
        \end{equation*}
        where $Z_i$ is the Pauli-Z operator acting on qubit $i$, and $\ket{\psi_{\text{out}}}$ is the final quantum state after applying the ansatz.
        This yields five output values in the range $[-1, 1]$ per convolution window.
        By applying this measurement procedure to all 169 overlapping windows across the input spectrum,
        we construct five output channels, each of length 169, resulting in an array of size $5\times 169$. The five output channels arise from measuring each of the $n_q=5$ qubits; they do not enforce or assume exactly five peaks. Finally, a ReLU activation function is applied to each output value, mapping the results to the range
        $[0, 1]$.
    }
    \item{
        \textbf{Max Pooling:} A max pooling layer with kernel size 5 (design choice to downsample $169 \rightarrow 33$) is applied to the quantum convolution output, reducing the array dimensions to $5\times 33$.
    }
    \item \textbf{Flattening:} The max pooled output is flattened into a one-dimensional array of size 165.
    \item{
        \textbf{Hidden Layer:} An intermediate fully connected layer with 128 neurons and ReLU activation
        processes the flattened output.
    }
    \item{
        \textbf{Output Layer:} The final output is generated by a fully connected layer with softmax activation,
        yielding an array of size 10. These five slots correspond to the dataset’s maximum of five peaks; if the task allowed more peaks, the slot count would be increased accordingly without changing the kernel size. To address the multi-task nature of our problem, this output array is logically divided into two components:
        the first five values represent a bitmask indicating occupied peak slots,
        while the remaining five values represent the corresponding peak positions per slot.
        This means we expect predictions of the form
        $\vec{y}' = (m_0, m_1, \ldots, m_4, p_0, p_1, \ldots, p_4)$,
        where $m_i \in \{0, 1\}$ indicates the presence of the $i$-th peak,
        and $p_i \in [0, 1]$ specifies its position.
        Since softmax outputs values in $[0, 1]$, the mask values $m_i \in [0,1]$ are thresholded at 0.5 to obtain binary presence indicators.
        Peak counting is then performed by summing the mask values: $\sum_{i=0}^{4} m_i$,
        while peak positions are extracted by applying the mask: $p'_i = m_i \cdot p_i$.
        
        This design choice allows the model to predict both the presence and position of peaks simultaneously.
        Specifically, the model always outputs five candidate peak positions $p_i$, but the binary mask $m_i$ determines
        whether each candidate is actually a peak ($m_i = 1$) or should be ignored ($m_i = 0$).
        This approach enables the model to handle variable numbers of peaks (0 to 5) within a fixed-size output structure,
        which simplifies the architecture and training procedure compared to variable-length output schemes.
    }
\end{enumerate}

Training this multi-task model requires a custom loss function that combines both task objectives.
Our loss function integrates binary cross-entropy loss for mask prediction with a permutation-invariant regression loss
for peak position estimation.
Specifically, we use the Hungarian loss~\cite{hungarian-method-1955}
to handle peak positions, which accounts for the fact that the order of the peaks does not matter.

Formally, let $\vec{y}$ denote the ground truth and $\vec{y}'$ the predicted output.
For simplicity, we decompose these into mask and position components:
$\vec{y}_m = (y_0, y_1, \ldots, y_4)$ and $\vec{y}_p = (y_5, y_6, \ldots, y_9)$ for the ground truth,
with corresponding predictions $\vec{y}'_m = (y'_0, y'_1, \ldots, y'_4)$ and $\vec{y}'_p = (y'_5, y'_6, \ldots, y'_9)$.
The combined loss function can then be written as:

\begin{equation}
    \begin{aligned}
        \mathcal{L}(\vec{y}, \vec{y}') =\ & \mathrm{BCE}(\vec{y}_m, \vec{y}'_m)\ + \\
        & \mathcal{H}(\vec{y}'_m \odot \vec{y}_p, \vec{y}'_m \odot \vec{y}'_p)
    \end{aligned}
    \label{eq:loss-function}
\end{equation}

where $\mathrm{BCE}$ is the binary cross-entropy loss, $\mathcal{H}$ is the Hungarian loss,
and $\odot$ is the element-wise multiplication operator. The Hungarian loss $\mathcal{H}$ solves an optimal assignment problem: it computes the minimal pairwise distance (e.g., MSE) between predicted positions $\vec{y}'_p$ and ground truth $\vec{y}_p$ across all possible peak permutations, ensuring order invariance \cite{hungarian-loss-2010}. While the mask contains only five binary values, BCE is appropriate here as it is the standard loss function for binary classification tasks,
even with small output spaces. This choice is consistent with common practice in multi-task learning frameworks
where binary classification components use BCE regardless of the number of classes \cite{gradient-based-learning-2002}.

The modular design of our quantum convolution layer enables the usage of different quantum circuits
within the same architectural model.

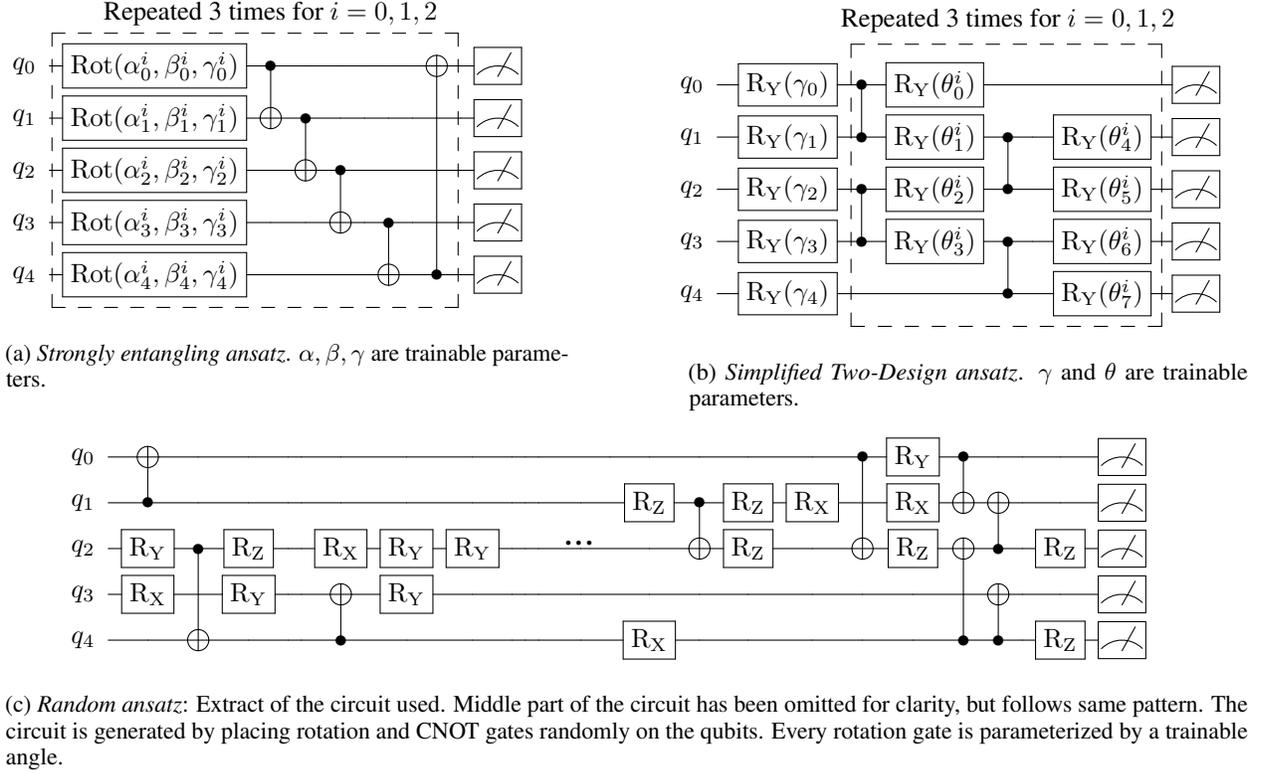
\begin{figure*}[h]
    \begin{subfigure}[t]{0.45\textwidth}
        \centering
        \[
            \Qcircuit @C=.5em @R=.3em @!R {
                & & \mbox{Repeated 3 times for $i=0,1,2$} & & & & & & & & \\
                \lstick{q_0} & \gate{\mathrm{Rot}(\alpha^i_0,\beta^i_0,\gamma^i_0)} & \ctrl{1} & \qw & \qw & \qw & \qw & \qw & \targ & \qw & \meter \\
                \lstick{q_1} & \gate{\mathrm{Rot}(\alpha^i_1,\beta^i_1,\gamma^i_1)} & \targ & \ctrl{1} & \qw & \qw & \qw & \qw & \qw & \qw & \meter \\
                \lstick{q_2} & \gate{\mathrm{Rot}(\alpha^i_2,\beta^i_2,\gamma^i_2)} & \qw & \targ & \ctrl{1} & \qw & \qw & \qw & \qw & \qw & \meter \\
                \lstick{q_3} & \gate{\mathrm{Rot}(\alpha^i_3,\beta^i_3,\gamma^i_3)} & \qw & \qw & \targ & \qw & \ctrl{1} & \qw & \qw & \qw & \meter \\
                \lstick{q_4} & \gate{\mathrm{Rot}(\alpha^i_4,\beta^i_4,\gamma^i_4)} & \qw & \qw & \qw & \qw & \targ & \qw & \ctrl{-4} & \qw & \meter
                \gategroup{2}{2}{6}{9}{.8em}{--}
            }
        \]
        \caption{\emph{Strongly entangling ansatz}. $\alpha, \beta, \gamma$ are trainable parameters.}
        \label{fig:strongly-entangling-ansatz}
    \end{subfigure}
    \hfill
    \begin{subfigure}[t]{0.45\textwidth}
        \centering
        \[
            \Qcircuit @C=.8em @R=.3em {
                & & & & \mbox{Repeated 3 times for $i=0,1,2$} & \push{\rule{0em}{2em}} & & \\
                & & & & & & & \\
                \lstick{q_0} & \gate{\mathrm{R_Y}(\mathrm{\gamma_0})} & \ctrl{1} & \gate{\mathrm{R_Y}(\theta^i_0)} & \qw & \qw & \qw & \meter \\
                \lstick{q_1} & \gate{\mathrm{R_Y}(\mathrm{\gamma_1})} & \control\qw & \gate{\mathrm{R_Y}(\theta^i_1)} & \ctrl{1} & \qw & \gate{\mathrm{R_Y}(\theta^i_4)} & \meter \\
                \lstick{q_2} & \gate{\mathrm{R_Y}(\mathrm{\gamma_2})} & \ctrl{1} & \gate{\mathrm{R_Y}(\theta^i_2)} & \control\qw & \qw & \gate{\mathrm{R_Y}(\theta^i_5)} & \meter \\
                \lstick{q_3} & \gate{\mathrm{R_Y}(\mathrm{\gamma_3})} & \control\qw & \gate{\mathrm{R_Y}(\theta^i_3)} & \ctrl{1} & \qw & \gate{\mathrm{R_Y}(\theta^i_6)} & \meter \\
                \lstick{q_4} & \gate{\mathrm{R_Y}(\mathrm{\gamma_4})} & \qw & \qw & \control\qw & \qw & \gate{\mathrm{R_Y}(\theta^i_7)} & \meter
                \gategroup{2}{3}{7}{7}{.8em}{--}
            }
        \]
        \caption{\emph{Simplified Two-Design ansatz}. $\gamma$ and $\theta$ are trainable parameters.}
        \label{fig:two-design-ansatz}
    \end{subfigure}%

    \begin{subfigure}[t]{\textwidth}
        \centering
        \[
            \Qcircuit @C=.5em @R=.3em @!R {
                \lstick{q_0} & \targ & \qw & \qw & \qw & \qw & \qw & \qw & \qw & \qw & \qw & \qw & \cds{4}{\textbf{\ldots}}
                & \qw & \qw & \qw & \qw & \ctrl{2} & \gate{\mathrm{R_Y}} & \ctrl{1} & \qw & \qw & \qw & \meter \\
                \lstick{q_1} & \ctrl{-1} & \qw & \qw & \qw & \qw & \qw & \qw & \qw & \qw & \qw & \qw & \qw
                & \gate{\mathrm{R_Z}} & \ctrl{1} & \gate{\mathrm{R_Z}} & \gate{\mathrm{R_X}} & \qw & \gate{\mathrm{R_X}} & \targ & \targ & \qw & \qw & \meter\\
                \lstick{q_2} & \gate{\mathrm{R_Y}} & \ctrl{2} & \gate{\mathrm{R_Z}} & \qw & \qw & \gate{\mathrm{R_X}} & \gate{\mathrm{R_Y}} & \gate{\mathrm{R_Y}} & \qw & \qw & \qw & \qw
                & \qw & \targ & \gate{\mathrm{R_Z}} & \qw & \targ & \gate{\mathrm{R_Z}} & \targ & \ctrl{-1} & \qw & \gate{\mathrm{R_Z}} & \meter\\
                \lstick{q_3} & \gate{\mathrm{R_X}} & \qw & \gate{\mathrm{R_Y}} & \qw & \qw & \targ & \gate{\mathrm{R_Y}} & \qw & \qw & \qw & \qw & \qw
                & \qw & \qw & \qw & \qw & \qw & \qw & \qw & \targ & \qw & \qw & \meter\\
                \lstick{q_4} & \qw & \targ & \qw & \qw & \qw & \ctrl{-1} & \qw & \qw & \qw & \qw & \qw & \qw
                & \gate{\mathrm{R_X}} & \qw & \qw & \qw & \qw & \qw & \ctrl{-2} & \ctrl{-1} & \qw & \gate{\mathrm{R_Z}} & \meter
            }
        \]
        \caption{
            \emph{Random ansatz}: Extract of the circuit used.
            Middle part of the circuit has been omitted for clarity, but follows same pattern.
            The circuit is generated by placing rotation and CNOT gates randomly on the qubits.
            Every rotation gate is parameterized by a trainable angle.
        }
        \label{fig:random-ansatz}
    \end{subfigure}
    \caption{Ansätze used in the quantum convolution layer.}
    \label{fig:quanv-ansatzes}
\end{figure*}

In this work, we implement and evaluate three distinct quantum circuits as ansätze, each exploring different aspects of
quantum computation.
These circuits, illustrated in~\autoref{fig:quanv-ansatzes}, are:

\begin{enumerate}[label=(\alph*)]
    \item{
        \emph{Strongly entangling ansatz}: This design maximizes quantum entanglement to leverage this uniquely
        quantum property.
        The circuit comprises three repetitions of parameterized rotation gates across all axes (Pauli-X, Y, and Z),
        followed by CNOT gate sequences that establish the entanglement.
    }
    \item{
        \emph{Simplified Two-Design ansatz (``Two-Design'')}: Inspired by the two-design architecture consisting of
        Pauli-Y rotations and controlled Z-entanglers proposed in~\cite{cost-function-dependent-barren-plateaus-2021}.
        It is a simplified version as it does not use universal 2-qubit gates.
        For simplicity, we will still refer to it as the Two-Design ansatz.
        The circuit begins with an initial layer of Pauli-Y rotations,
        followed by three repetitions combining controlled Z-gates with additional Pauli-Y rotations.
    }
    \item{
        \emph{Random ansatz}: Based on the original design from~\cite{quanvolutional-neural-networks-2020},
        this circuit features a randomly generated arrangement of rotation and CNOT gates.
        Unlike the original implementation, where rotation parameters were frozen during training,
        our version treats all rotation gates as trainable parameters.
    }
\end{enumerate}

\subsection{Mathematical Analysis of Quantum Kernel Performance}

This section provides a theoretical analysis of the quantum kernels based on expressibility, entanglement capability, and trainability principles.
The analysis is grounded in established quantum information theory, though the empirical validation of these theoretical predictions is presented in Section~\ref{sec:experiments-and-results}.
Let us analyze each kernel from first principles.

\subsubsection{Strongly Entangling Ansatz}
The superior performance of the strongly entangling ansatz stems from its high expressibility and entanglement generation. Mathematically, the circuit implements a unitary transformation $U(\vec{\theta})$ that efficiently explores the unitary group $\mathrm{SU}(2^n)$ compared to circuits with limited entanglement connectivity. 
The sequential application of parameterized rotations and CNOT gates creates a rich family of states:

\begin{equation}
    \ket{\psi_{\text{out}}} = \prod_{l=1}^{L}\left[\left(\bigotimes_{i=1}^{n} R_{XYZ}(\theta_i^l)\right) \cdot \mathrm{CNOT}_{\text{chain}}\right] \ket{\psi_{\text{in}}}
\end{equation}

where $L$ denotes the number of layers (circuit depth), $n = 5$ is the number of qubits,
$R_{XYZ}(\theta_i^l)$ represents parameterized rotations around the X, Y, and Z axes for qubit $i$ in layer $l$,
and $\mathrm{CNOT}_{\text{chain}}$ denotes the chain of CNOT gates that create entanglement between adjacent qubits.
The entanglement entropy $S(\rho_A) = -\mathrm{Tr}(\rho_A \log \rho_A)$ for bipartitions of the system grows rapidly with circuit depth $L$, enabling the capture of complex spectral correlations.
Here, $\rho_A = \mathrm{Tr}_B(\rho)$ is the reduced density matrix obtained by tracing out subsystem $B$ from the full system density matrix $\rho = \ket{\psi_{\text{out}}}\bra{\psi_{\text{out}}}$,
and a bipartition refers to dividing the $n$-qubit system into two complementary subsystems $A$ and $B$ (e.g., qubits $\{0,1,2\}$ and $\{3,4\}$).
This is particularly crucial for separating overlapping peaks where classical methods fail due to their limited feature space.

\subsubsection{Simplified Two-Design Ansatz}
We include the Two-Design ansatz in our comparison to evaluate how different entanglement structures affect performance, even though theoretical considerations suggest it may be less expressive.
The suboptimal performance of the Two-Design ansatz arises from its restricted entanglement structure.
While it maintains trainability by avoiding barren plateaus \cite{cost-function-dependent-barren-plateaus-2021},
the limited entanglement generation constrains its ability to represent complex spectral features.
Specifically, the Two-Design ansatz creates only partial (weak) entanglement rather than strong, long-range entanglement,
which restricts its expressibility compared to the strongly entangling ansatz~\cite{expressibility-2019}:

\begin{equation}
    \mathcal{E}_{\text{Two-Design}} \ll \mathcal{E}_{\text{Strongly-Entangling}}
\end{equation}

where $\mathcal{E}$ quantifies the achievable entanglement (e.g., measured via the average entanglement entropy across all bipartitions).
The absence of multi-qubit correlations beyond nearest neighbors limits the kernel's capacity to capture the long-range dependencies present in overlapping spectral features. 
This expressibility limitation can be understood through the lens of quantum circuit expressibility \cite{expressibility-2019}: circuits with limited entanglement connectivity explore a smaller subset of the unitary group $\mathrm{SU}(2^n)$, reducing their ability to represent complex feature mappings compared to circuits with full entanglement connectivity.

\subsubsection{Random Ansatz}
The inconsistent performance of the random ansatz illustrates the barren plateau phenomenon. For random parameter initializations, the variance of gradients vanishes exponentially with system size \cite{barren-plateaus-2018}:

\begin{equation}
    \mathrm{Var}[\partial_{\theta_i} \mathcal{L}] \in \mathcal{O}\left(\frac{1}{2^n}\right)
\end{equation}

This gradient vanishing makes optimization practically impossible for larger systems, explaining the high variance in results across different random initializations.

In summary, based on theoretical analysis, the strongly entangling ansatz is expected to demonstrate superior performance due to its ability to create highly entangled states that capture complex correlations in spectral data. Mathematically, this can be understood through the expressibility of the ansatz: the sequential application of parameterized rotations and controlled gates creates a rich family of unitary transformations that effectively map local spectral features to globally relevant patterns. 

In contrast, the simplified Two-Design ansatz, while computationally efficient, exhibits limited entanglement capabilities, restricting its ability to capture the complex interdependencies between overlapping peaks. The random ansatz suffers from trainability issues due to the barren plateau phenomenon, where gradients vanish exponentially with system size, explaining its inconsistent performance across different runs. The empirical results presented in Section~\ref{sec:experiments-and-results} confirm these theoretical predictions.

\subsubsection{Model Training}
The model is trained using the Adam optimizer~\cite{adam-optimization-2017} with a learning rate adjusted by
a cosine annealing scheduler, which was first proposed in~\cite{stochastic-gradient-descent-with-warm-restarts-2017},
with a starting learning rate of $0.01$ and a minimum learning rate of $10^{-4}$.
Additionally, a dropout layer is inserted during training between the hidden layer and the output layer with a 10\% dropout chance
to avoid overfitting.
The model is trained for 100 epochs with a batch size of 16.

The circuits and the gradient descent were implemented using the PennyLane library~\cite{pennylane-2022},
which provides seamless quantum backend switching, has automatic differentiation capabilities, and
integrates well with PyTorch~\cite{pytorch2-2024}, used for the classical parts of the model.

\subsection{Classical Convolutional Neural Networks}\label{subsec:classical-convolutional-neural-networks}

The classical CNN used in this work is a direct counterpart to the QuanvNN\@.
This means that the architecture is the same, except for the quantum convolution layer,
which is replaced by a classical one-dimensional convolution layer
(i.e., step (B) in~\autoref{fig:quanv-architecture} is replaced by a classical convolution layer).
The convolution layer is configured like the quantum convolution layer, with a kernel size of 32,
a stride size of one, no padding, and five output channels.
Training procedures, including optimizer, learning rate scheduling, and dropout, remain consistent across both architectures
to ensure fair comparison.
We adopt this architectural equivalence strategy to isolate the contribution of the quantum convolution layer from other architectural choices.
By maintaining identical network depth, layer sizes, and training hyperparameters, any performance differences can be attributed to the quantum versus classical nature of the convolution operation itself, rather than differences in model capacity or training methodology.
This approach follows the standard practice in quantum machine learning literature for comparing quantum and classical models \cite{quanvolutional-neural-networks-2020}.
However, we acknowledge that an identical architecture may not necessarily represent the optimal classical baseline, as classical CNNs might benefit from different architectural choices (e.g., different kernel sizes, additional layers, or alternative activation functions) that are not directly applicable to quantum circuits.
The goal of this comparison is not to demonstrate quantum advantage in absolute terms, but rather to provide a controlled comparison that highlights the specific characteristics and trade-offs of quantum convolution operations within the same architectural framework.

    \section{Experiments and Results}\label{sec:experiments-and-results}
    We evaluate the approaches from~\autoref{sec:approach} for peak finding on a simplified NMR-inspired dataset,
focusing on the feasibility of using current quantum computing technology for this real-world problem. While our dataset is simplified compared to full experimental NMR spectra, the fundamental challenges of peak finding - overlapping peaks, noise, and variable peak counts - are preserved and systematically controlled.
This allows us to isolate and study the specific advantages of quantum approaches while maintaining generalizability to real-world scenarios, as the core mathematical structure of the peak detection problem remains consistent regardless of dataset complexity.
We address the following research questions:

\begin{itemize}
    \item How do QuanvNNs compare to classical CNNs in terms of convergence, training stability, and error rates?
    \item Given the expressiveness of the high-dimensional Hilbert space available to quantum circuits and given the still limited amount of qubits in present hardware, can we obtain indications that QuanvNNs can outperform classical CNNs on relatively harder problems within current hardware constraints??
    \item What is the impact of noise on the performance of QuanvNNs compared to classical CNNs?
    Can QuanvNNs maintain their performance under noisy conditions, which are common in quantum computing?
\end{itemize}

\subsection{Datasets}\label{subsec:datasets}
We created two datasets of spectra for training and validation: a mixed dataset containing easy, medium, and hard spectra, and a hard dataset containing only hard spectra.

Peak finding is a common problem in spectrum analysis, such as NMR or mass spectrometry. Since the focus of this work is methodological, we use synthetic data; a common practice in the development of methods for spectrum analysis, as experimental data is less readily available (\cite{zipoli2025ir}).
Hand-crafting a dataset provides the advantage of full control over problem difficulty and parameters,
including peak count, noise levels, peak widths, and overlaps.
For this work, we created two datasets of synthetic NMR-inspired spectra: a mixed dataset with varying
difficulty levels (1150 spectra total) and a hard dataset containing only complex spectra (1000 spectra total).
Within the mixed dataset, we have three difficulty levels: \emph{easy}, \emph{medium}, and \emph{hard}.
The easy spectra consist of a small number of peaks with low noise and low overlap probabilities (using a distribution with $\mu = 1$ and $\sigma = 0.96$ that concentrates peak placements toward section centers, resulting in approximately 15-20\% probability of peak overlap).
The medium spectra have up to five peaks, with moderate noise and the same overlap probabilities as the easy spectra.
Hard spectra also contain up to five peaks, but with high noise, wider peaks, and high overlap probabilities.

\begin{figure}[h]
    \centering
    \includegraphics[width=0.85\linewidth]{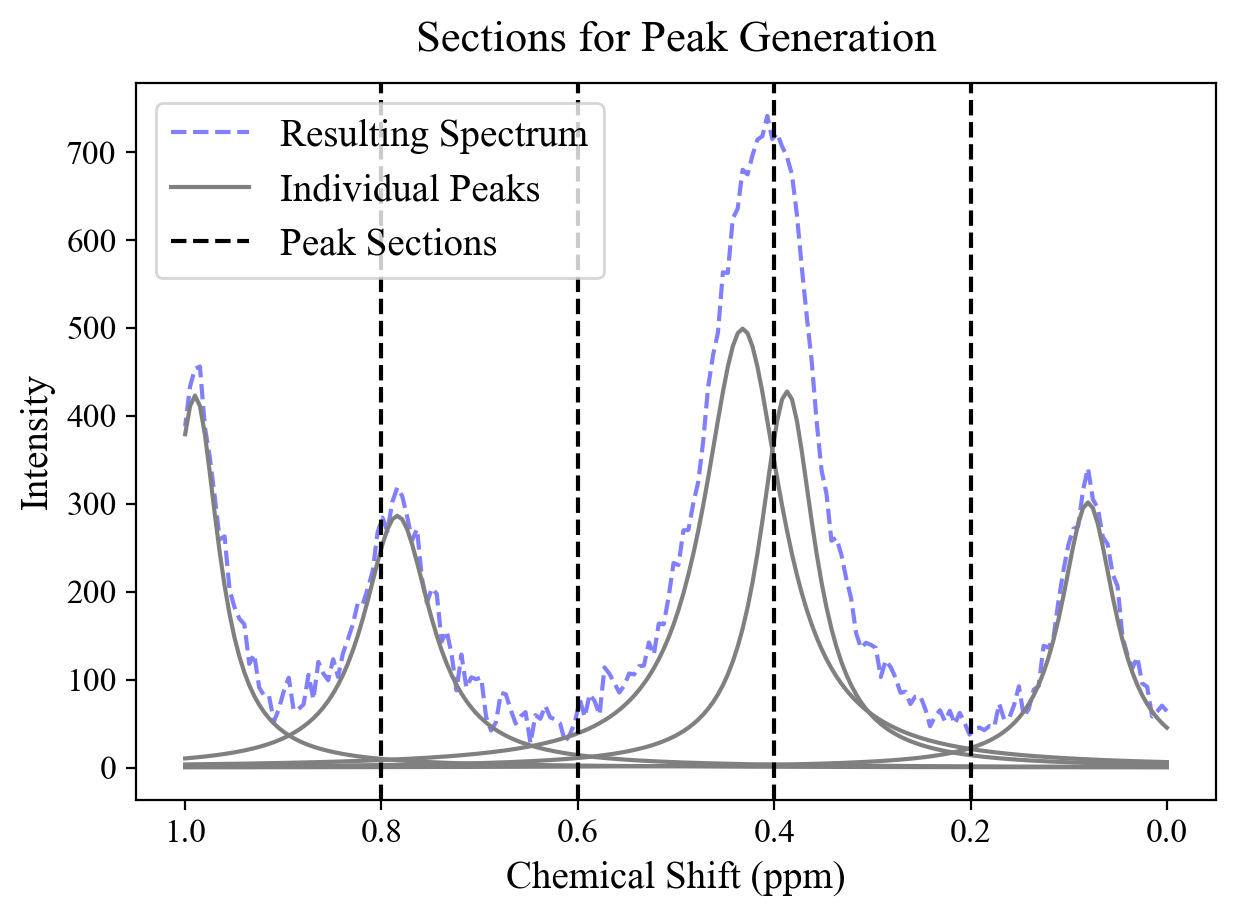}
    \caption{
        The peak placement sections of this five-peak example spectrum visualized:
        To avoid an uneven distribution of peaks and be in control of overlap probabilities,
        the peak placement is ensured by subdividing the spectrum into sections.
        In this example, the spectrum is divided into five sections in which a peak is placed.
        Furthermore, as it is a hard spectrum, the peak positions are concentrated towards the edges of the sections,
        allowing for more peaks to be placed closer together and thus creating more overlaps.
    }
    \label{fig:peak-sections}
\end{figure}

To control problem difficulty, we designed configurable parameters for the dataset generator,
including peak count range per spectrum, SNR, peak width range, intensity range,
and overlap probabilities.
The spectrum generator operates as follows:
First, it chooses how many peaks will be present in the spectrum, using the specified peak count range.
The spectrum is then subdivided into as many sections as there are peaks to ensure even distribution.
An example of this process for five peaks is shown in~\autoref{fig:peak-sections}.
Within each section, a peak is placed, with the peak position being randomly chosen using a
customizable probability distribution, which we describe in detail later.
For every peak placed, a peak width and intensity are randomly chosen within the specified ranges.
With all peaks specified, the corresponding Lorentzian function is calculated per peak.
The resulting spectrum is then generated by summing Lorentzian peak functions \cite{nmr-lorentzian-peaks-2023}:
\[ I(\nu) = \sum_k \frac{A_k}{1 + [2(\nu - \nu_k)/\Gamma_k]^2} \]
where $A_k$, $\nu_k$, and $\Gamma_k$ denote amplitude, position, and linewidth of the $k$-th peak.
Finally, the generator adds Gaussian noise to the spectrum, with a specified SNR\@.

As previously mentioned, peak positions within sections are not uniformly distributed.
Instead, we use a custom probability distribution function that satisfies the following requirements:

\begin{itemize}
    \item With higher overlap probabilities, the distribution function should be concentrated towards the edges of
    the sections, allowing for more peaks to be placed closer together.
    \item With lower overlap probabilities, the distribution function should be more concentrated towards the center of the section,
    allowing for more evenly distributed peaks.
    \item The probability at the exact edges of the section should be very low, to avoid peaks being placed on the
    edges themselves, which could lead to peaks being placed on top of each other.
\end{itemize}

We define a function that combines two log-normal distributions: one for the left edge and one for the right
edge of each section.
Formally, let $F(x)$ be the log-normal probability density function defined as:

\begin{equation}
    F(x) = \frac{1}{x \sigma \sqrt{2\pi}} \exp{-\frac{(\ln x - \mu)^2}{2\sigma^2}}
    \label{eq:log-normal-pdf}
\end{equation}

where $\mu$ is the mean and $\sigma$ is the standard deviation, both adjustable to control the distribution spread.
Then, our custom probability distribution function $P(x)$ for the peak positions is defined as:

\begin{equation}
    P(x) = \frac{F(5x) + F(5 - 5x)}{\int_0^1 \left[ F(5x) + F(5 - 5x) \right] \, dx}
    \label{eq:custom-pdf}
\end{equation}

\begin{figure}[h]
    \centering
    \includegraphics[width=\linewidth]{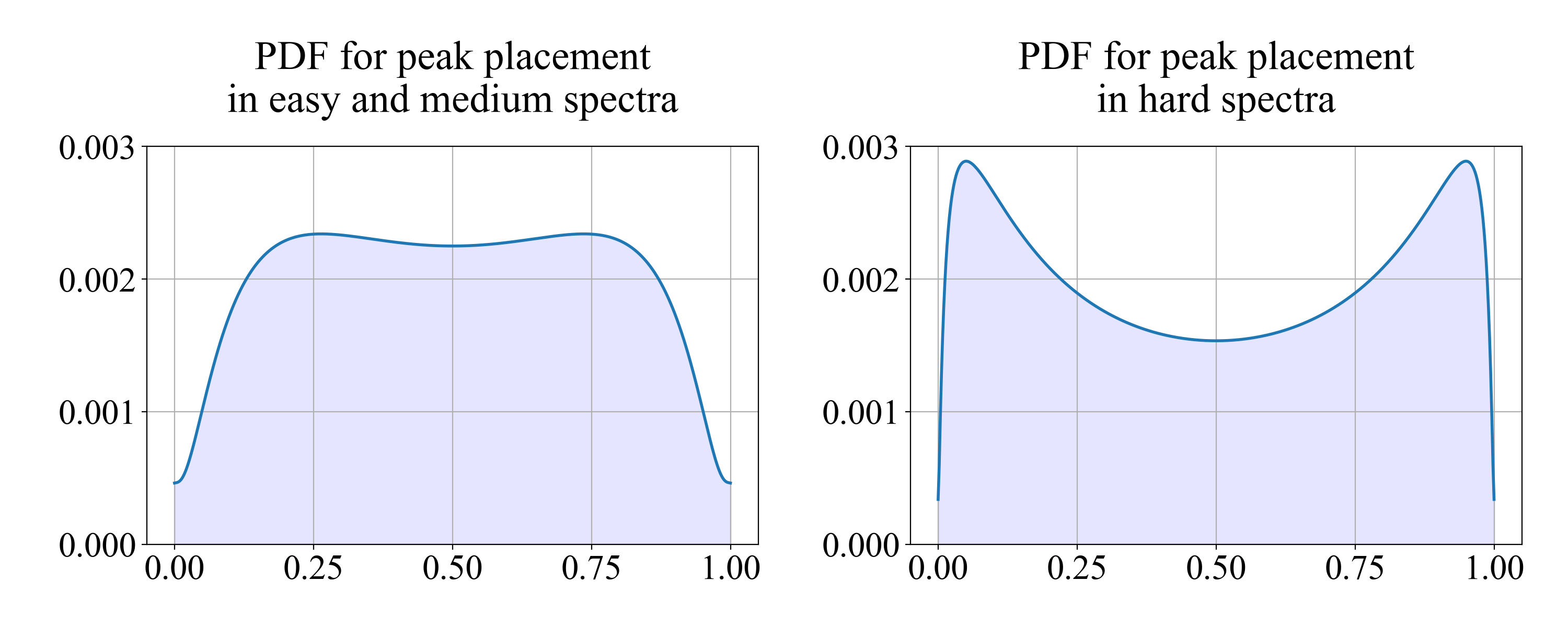}
    \caption{
        The custom probability distribution functions used for peak placement.
        The easy and medium spectra use the left distribution that is configured with $\mu = 1$ and $\sigma = 0.96$.
        Hard spectra use the right distribution that is configured with $\mu = 0.85$ and $\sigma = 1.5$.
    }
    \label{fig:double-log-normal-distribution}
\end{figure}

By adjusting the parameters $\mu$ and $\sigma$ of these log-normal distributions, we control the distribution shape
and thus the peak placement probabilities.
The scaling factor of 5 in the log-normal function argument was chosen empirically
to achieve the desired peak placement behavior.
The resulting distributions used for the different spectra
difficulties are shown in~\autoref{fig:double-log-normal-distribution}.

\begin{figure*}[h]
    \centering
    \includegraphics[width=\linewidth]{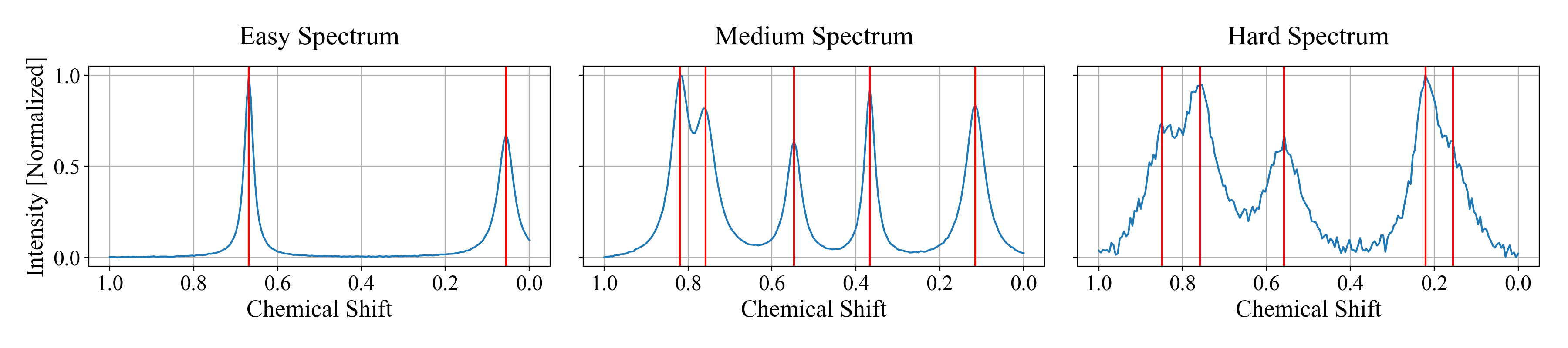}
    \caption{
        Example spectra for each difficulty configuration.
    }
    \label{fig:example-spectra}
\end{figure*}

Overall, we used the following parameters for the different spectra difficulties:

\begin{itemize}
    \item \textbf{Easy spectra:} Contain zero to two peaks, with a linewidth range of $[0.02, 0.1)$, and an
    intensity range of $[500, 1000)$.
    Peak positioning is done using the left distribution in~\autoref{fig:double-log-normal-distribution} with $\mu = 1$ and $\sigma = 0.96$.
    A low Gaussian noise with magnitude of 1 is added to the spectrum.
    \item \textbf{Medium spectra:} Contain three to five peaks, with a linewidth range of $[0.025, 0.12)$, and an
    intensity range of $[500, 1000)$.
    Peak positioning is also done using the left distribution in~\autoref{fig:double-log-normal-distribution} with $\mu = 1$ and $\sigma = 0.96$.
    A Gaussian noise with magnitude of 2 is added to the spectrum.
    \item \textbf{Hard spectra:} Contain three to five peaks, with a linewidth range of $[0.1, 0.22)$, and an intensity
    range of $[100, 500)$, making the peaks appear closer together.
    Peak positioning is done using the right distribution in~\autoref{fig:double-log-normal-distribution} with $\mu = 0.85$ and $\sigma = 1.5$.
    A Gaussian noise with magnitude of 14 is added to the spectrum.
\end{itemize}

Example spectra for each difficulty configuration are shown in~\autoref{fig:example-spectra}.

\subsubsection{Mixed Dataset}\label{subsubsec:mixed-dataset}

\begin{figure}[h]
    \centering
    \includegraphics[width=0.75\linewidth]{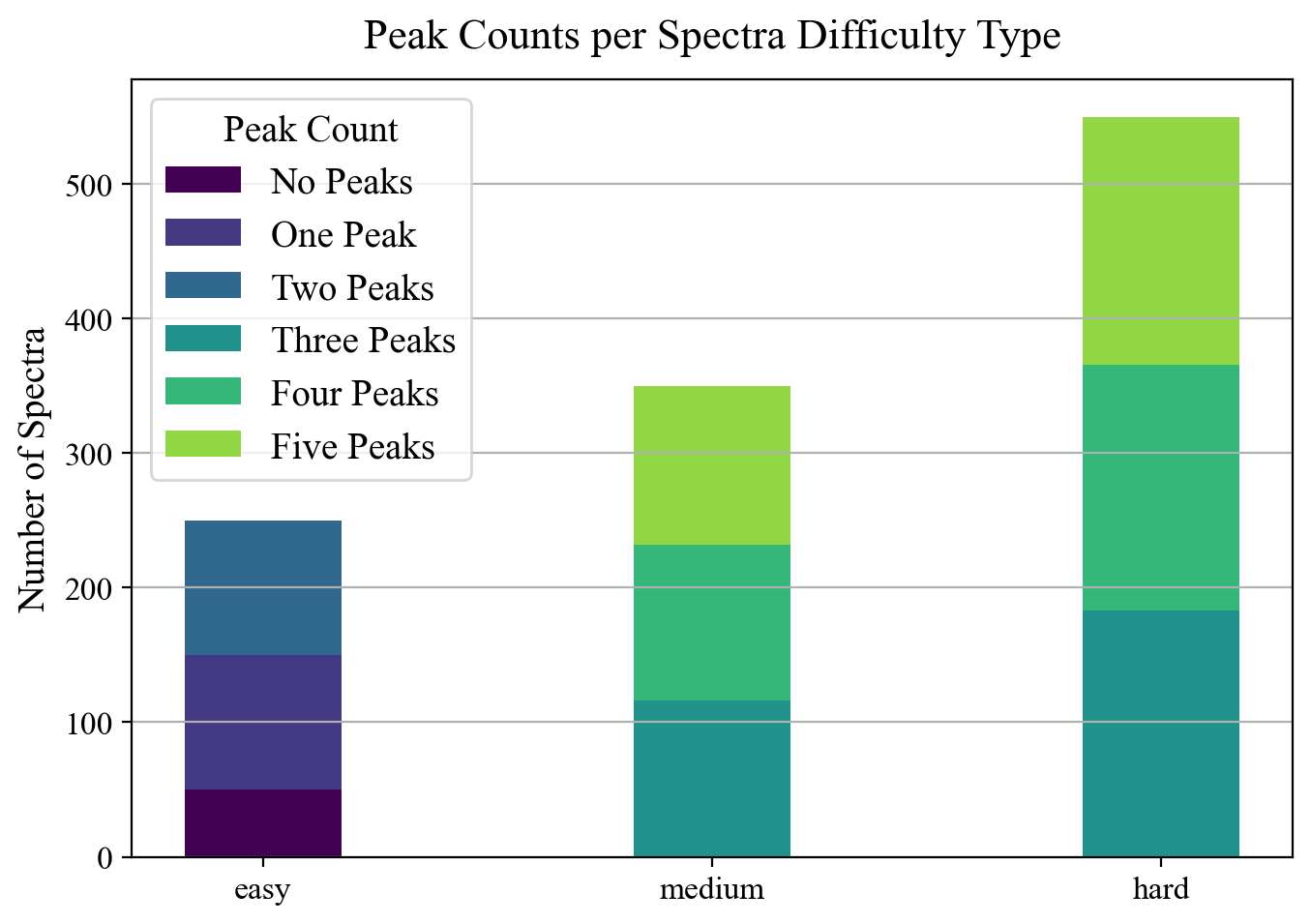}
    \caption{ The distribution of the number of peaks per spectrum difficulty in the mixed dataset. }
    \label{fig:peak-counts-per-spectra-type-mixed-dataset}
\end{figure}

The mixed dataset comprises 1150 synthetic spectra: 550 generated using hard parameters,
350 using medium parameters, and 250 using easy parameters.
Additionally, we restrict the easy spectra generator to produce only 50 spectra with no peaks.
The resulting distribution of the number of peaks per spectrum difficulty
is shown in~\autoref{fig:peak-counts-per-spectra-type-mixed-dataset}.
The dataset is imbalanced towards the challenging spectra as these are the most interesting cases, and
we also expect the models to easily adapt to the easier spectra.

A train/validation/test split of almost 80/10/10\% is applied to the dataset, resulting in 919 training spectra, 116
validation spectra, and 115 test spectra.
We apply stratified splitting to the subsets, preserving both the peak count distribution per spectrum and
the difficulty level distribution across all subsets (which is the reason why the split is not exactly 80/10/10\%,
but one off).
While the test set spectra remain fixed across all models, the training and
validation sets are reshuffled before each training run according to the run-specific random seed.

\subsubsection{Hard Dataset}\label{subsubsec:hard-dataset}

The hard dataset consists of 1000 spectra, all generated using the hard parameter settings described in the previous
section (example spectrum shown in~\autoref{fig:example-spectra}, right side).
These parameters create increasingly challenging conditions:

\begin{itemize}
    \item \textbf{High peak overlap}: Wider linewidths ($[0.1, 0.22)$) and edge-concentrated placements (\autoref{fig:double-log-normal-distribution}, right) increase peak interactions
    \item \textbf{Low signal-to-noise ratio (SNR)}: High noise magnitude (14) obscures low-intensity peaks ($[100, 500)$)
    \item \textbf{Increased ambiguity}: Broad peaks and frequent overlaps exacerbate the ill-posed inverse problem (Fig.~\ref{fig:ill-posedness})
\end{itemize}

A train/validation/test split of 80/10/10\% is applied to the dataset, resulting in 800 training spectra, 100
validation spectra, and 100 test spectra.

\subsection{Results}\label{subsec:results}

We evaluate all models on both datasets using F1 score, recall, and precision metrics for peak counting,
and mean absolute error (MAE) and mean squared error (MSE) for peak position estimation.
For the multi-class peak counting task (six classes for 0 to 5 peaks),
we use weighted averages of F1 score, recall, and precision across all classes.
To ensure a statistically significant evaluation, we train each model with different random seeds and report the mean
and standard deviation of the results.

Model training and evaluation utilized Pennylane's \texttt{lightning.qubit}
simulator backend for both ideal and noisy quantum models. For ideal quantum circuits, gradient computation uses the parameter-shift rule (see Section~\ref{sec:fundamentals-of-quantum-computing}) or adjoint differentiation~\cite{efficient-calculation-of-gradients-2020}.
Noise simulation made use of Pennylane's noise model, created from Qiskit's generic backend noise characteristics.
The generic Qiskit backend is initialized with error data generated and sampled randomly from historical IBM backend data.
We approximate readout errors with amplitude damping since they dominate NISQ device infidelity~\cite{ibm-quantum-errors-2023}.
As Pennylane's noise model does not support classical readout errors, we compensate for this by adding an amplitude
damping error channel to every qubit upon measurement.

Introducing noise to quantum models necessitates changing the gradient descent technique, as the
adjoint method and parameter-shift rule only work for ideal quantum circuits.
Consequently, for noisy quantum models, we used the Simultaneous Perturbation Stochastic Approximation
(SPSA) algorithm~\cite{adaptive-stochastic-approximation-2000} instead of gradient-based optimization.
Additionally, since adding noise increases the computational cost of evaluating these models significantly, we
evaluated the first 80 epochs on the ideal simulator and then switched to the noisy simulator for the remaining epochs.

An overview of the results on the mixed dataset is shown in~\autoref{tab:results-table}
(with the results only for the hard subset of the mixed dataset in Table~\ref{tab:results-table}b)
and on the hard dataset in~\autoref{tab:hard-results-table}.

\subsubsection{Discussion of Mixed Dataset Results}\label{subsubsec:discussion-mixed-dataset}

\begin{figure*}[h]
    \centering
    \includegraphics[width=0.96\linewidth]{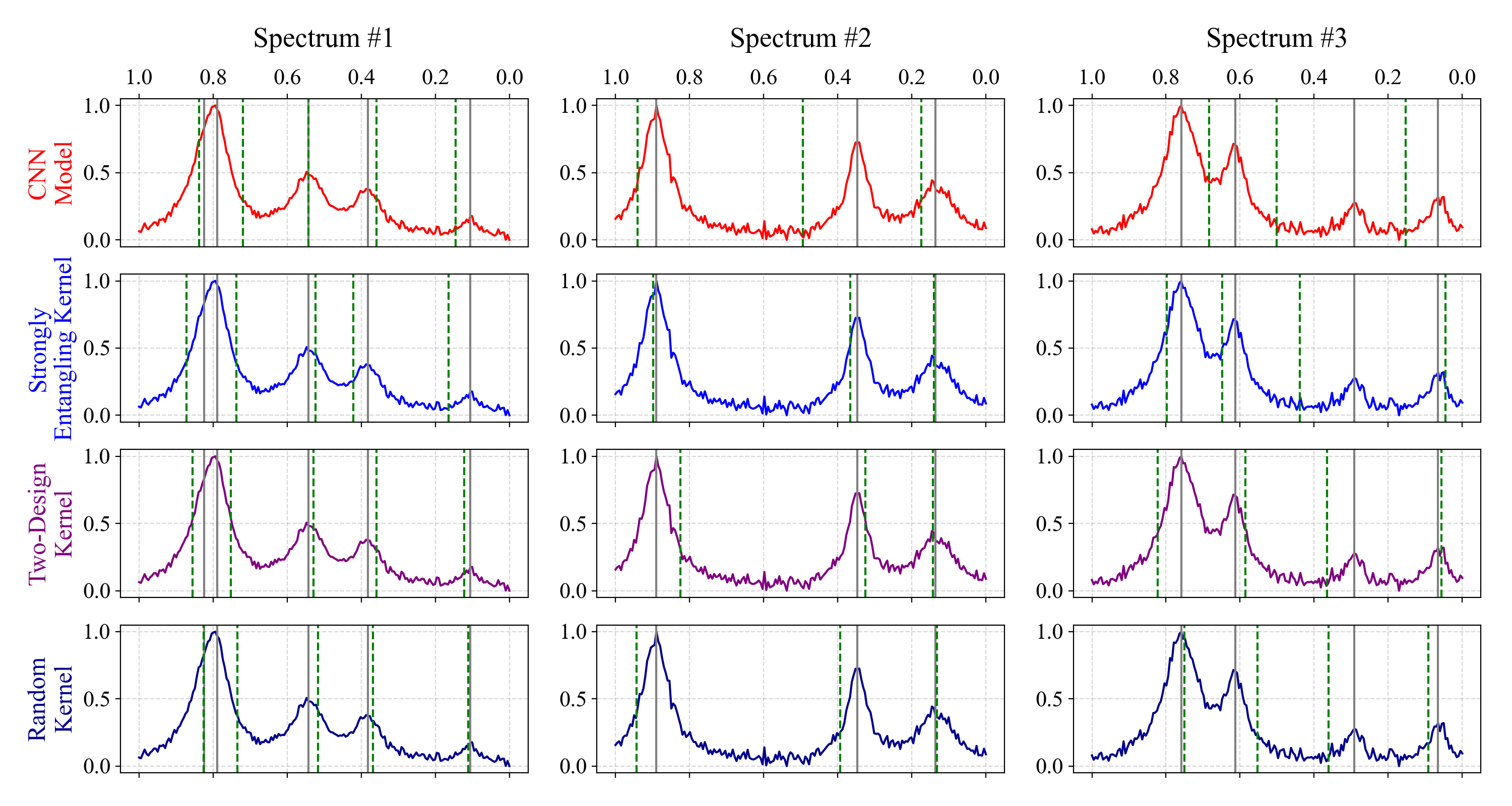}
    \caption{
        Comparison of the predictions of the ideally simulated models and the 
        classical convolutional model on the \emph{mixed dataset}.
        The best performing run per model (based on F1 and MAE score) was chosen to generate the predictions.
        The \textcolor{gray}{gray line} indicates the actual peak position,
        while the \textcolor{darkgreen}{dashed green line} indicates the predicted peak position.
        The three spectra were manually chosen as they reveal different characteristics of the models.
    }
    \label{fig:mixed-predictions-comparison}
\end{figure*}

The results on the mixed dataset appear remarkably similar across all difficulty levels and models.
Although there are small differences in F1 scores for peak counting and MAE for peak position estimation,
only the random kernel model stands out with a 1.55\% higher F1 score than the classical convolutional model.
To check whether the differences between the quantum models and the classical model are statistical anomalies, we
performed a one-sided Wilcoxon signed-rank test on the F1 scores and mean absolute errors of the quantum models and
the classical convolutional model.
The null hypothesis states that there is no difference in performance between the models, while the alternative hypothesis
states that the quantum models perform better than the classical model (i.e., the F1 score is higher and the MAE is lower).
Only the random kernel model shows a statistically significant difference in F1 score
with a $p$-value of $0.0101$ and in MAE with a $p$-value of $0.0015$.
Moreover, noise appears to generally have a negative impact on the performance of the quantum models.
All the noisy models perform worse on average than the ideally simulated models.

However, this performance degradation becomes particularly relevant when examining the performance across individual difficulty levels, as discussed below.
Filtering the predictions on the test set by difficulty level reveals that the models perform similarly on the easy and
medium spectra.
However, we can see a difference in performance when only looking at the hard
spectra, as shown in Table~\ref{tab:results-table}b.
We observe that the classical convolutional model now performs consistently worse in every metric than the quantum
models.
Especially for the peak counting task, all ideally simulated quantum models outperform the classical model by about 3\%
in F1 score.
The noisy quantum models also outperform the classical model, but the difference is smaller, with an improvement of
about 1\% in F1 score.
Differences can also be noticed when we look at the resulting predictions of the models on the hard subset of the
mixed dataset, as shown in~\autoref{fig:mixed-predictions-comparison}.
The figure shows the predictions of the best performing run for each respective model on three different spectra.
The first spectrum presents an interesting case, containing a challenging overlap of two peaks that could be
interpreted as a single peak.
Although all models correctly identified both peaks, their approaches differed.
The classical model and the random kernel model predicted the first peak position correctly, but the second peak
position is shifted considerably upfield (i.e., lower frequency).
For the second spectrum, the classical model exhibits an unusual misplacement of the middle peak despite it being
well-defined, while the quantum models show less dramatic errors.
Lastly, with the third spectrum, we see that the classical model did various mistakes:
It guessed the wrong number of peaks and position estimations are off by a large margin.
However, quantum models correctly identified the number of peaks.
Position estimation is also slightly off, depending on the model, but the predictions are still closer to the actual
peak positions than the classical model.
Another notable observation is that all models struggle with the peak at around 0.3 ppm, which is a very low
intensity peak.
All predictions are off by a large margin; we can only see slight improvements for the quantum models,
hinting that both classical and quantum models repeat similar mistakes.
However, given that the models are structurally the same except for the convolution, this was expected.

\subsubsection{Discussion of Hard Dataset Results}\label{subsubsec:discussion-hard-dataset}

Results on the hard dataset emphasize the trends observed in the mixed dataset's hard subset.
Here, quantum models demonstrate clear superiority over the classical model across all metrics.
For the strongly entangling kernel model, we observe an average improvement of 10.9\% in F1 score and a 29.8\%
improvement in MAE for peak position estimation.

\begin{figure}[ht]
    \centering
    \begin{subfigure}[b]{0.85\linewidth}
        \centering
        \includegraphics[width=\linewidth]{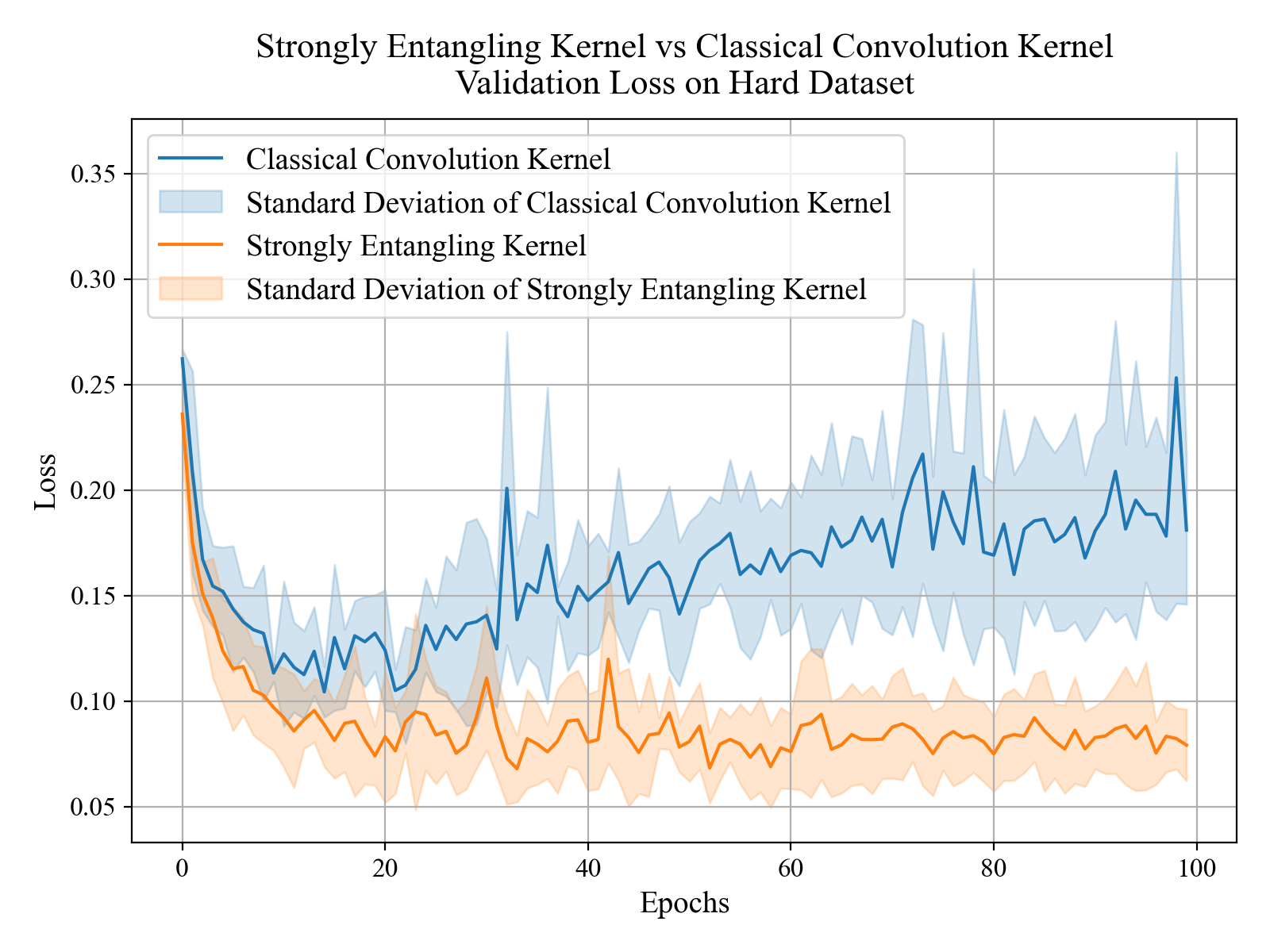}
        \caption{ \emph{Validation loss} }
        \label{fig:hard-validation-loss}
    \end{subfigure}
    \begin{subfigure}[b]{0.85\linewidth}
        \centering
        \includegraphics[width=\linewidth]{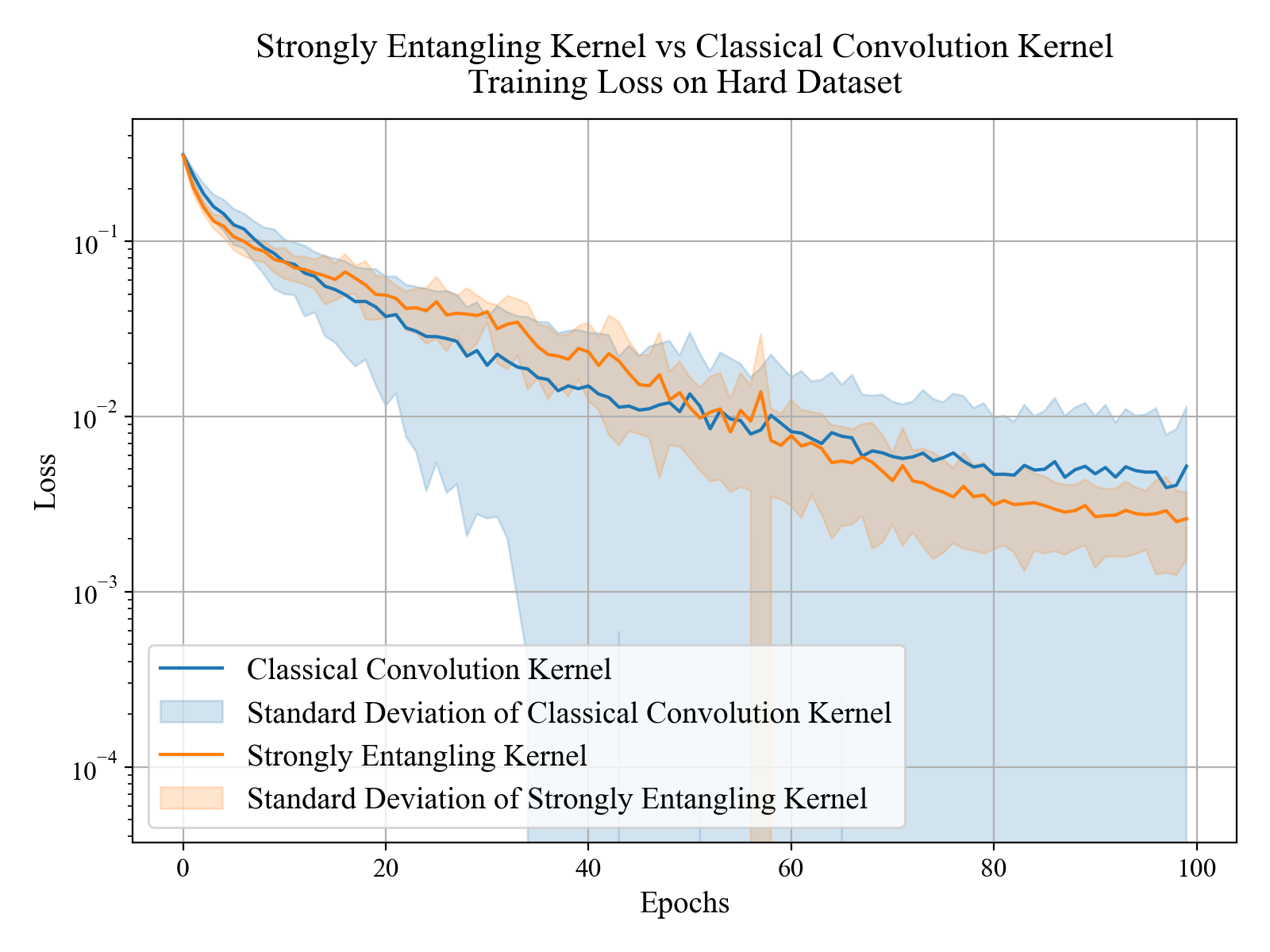}
        \caption{
            \emph{Training loss}.
            Y-axis is on a logarithmic scale to better visualize the differences.
        }
        \label{fig:hard-train-loss}
    \end{subfigure}
    \caption{
        Loss curves of the best performing quantum model (strongly entangling kernel) and the classical
        convolutional model on the \emph{hard dataset} over the training epochs.
    }
    \label{fig:hard-loss-curves}
\end{figure}

Beyond performance improvements, we observe significant differences in convergence behavior,
as illustrated in~\autoref{fig:hard-loss-curves}.
During training, the quantum models converge more stably than the classical model, meaning that the initial
weights do not have as much impact on the final performance.
Across all runs, the classical models show higher variance in training and validation loss consistently over every epoch. A concerning observation is that the classical model shows validation loss divergence after around 40 epochs, while the training loss continues to decrease.
This pattern - decreasing training loss accompanied by increasing validation loss - indicates that the classical model is overfitting and that our training approach (learning rate, regularization, or early stopping strategy) may not be optimal for the classical architecture.
In contrast, the quantum models continue to converge on both training and validation sets, suggesting that the quantum convolution layer provides better regularization or that the training dynamics are more favorable for the quantum models under the current training configuration.

Furthermore, we observe again that noise has a negative impact on the overall performance of the quantum models,
although the difference is negligible for the two-design kernel model.
The noisy strongly entangling kernel model, however, suffers from a significant performance drop in both F1 score
and MAE, despite its noise-free counterpart showing the best performance.

\subsubsection{Discussion of Trainable Parameters}\label{subsubsec:discussion-trainable-parameters}

\begin{figure}[h]
    \centering
    \includegraphics[width=0.85\linewidth]{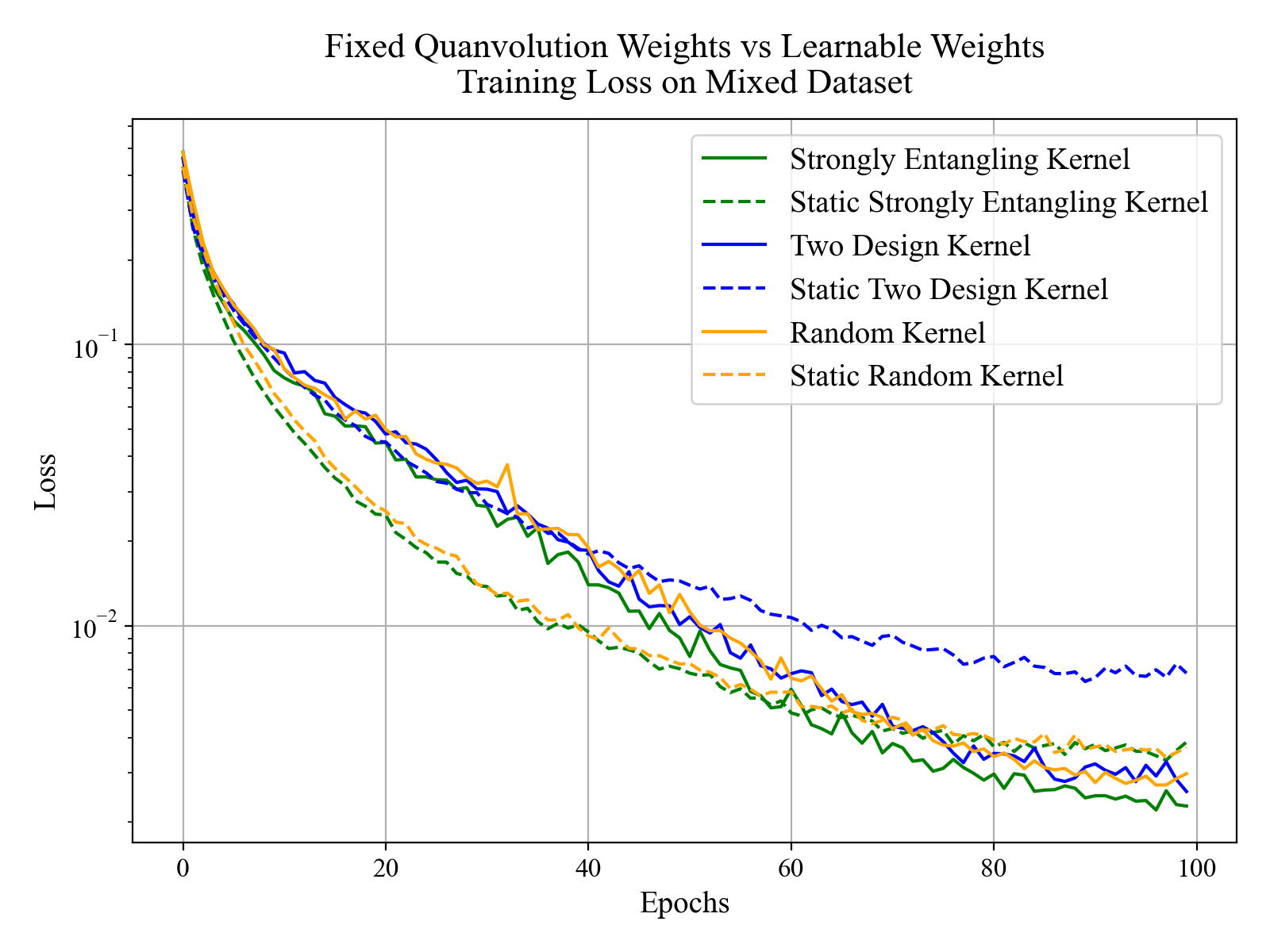}
    \caption{
        Comparison of training loss on the mixed dataset for static versus learnable quanvolutional layers.
        Static layers have weights initialized randomly and frozen during training,
        while learnable layers have weights optimized during training (the approach used in this work).
    }
    \label{fig:mixed-static-learnable-train-loss}
\end{figure}

Our implementation employs trainable parameters in the quanvolutional layer,
contrasting with related work~\cite{quanv4eo-2025, quanvolutional-neural-networks-2020} that typically uses a lazy training approach where
quanvolutional layer parameters remain fixed during training.
To evaluate the impact of our learnable approach, we trained our three models with the same randomly initialized
quanvolutional layer parameters, but without optimizing them during training.
The comparison of this lazy training approach with our learnable
approach is shown in~\autoref{fig:mixed-static-learnable-train-loss}.
We notice that for every model, the training loss eventually becomes lower for the learnable quanvolutional layer
than for the static quanvolutional layer. Quantitatively, the learnable approach achieves on average 15-25\% lower final training loss compared to the static approach across all three quantum models, indicating that the optimization of the parameters is beneficial for the
model performance.
The speed with which the learnable models reach lower training loss is not consistent across the models.
While the two-design kernel model already reaches lower training loss after 40 epochs, the strongly entangling kernel
and random kernel models reach a lower training loss after around 60 to 70 epochs respectively.
When considering the computational cost of optimizing the parameters, we observe that training the parameters of the
quanvolutional layer may also be neglected for some model configurations, especially for large models.

\begin{table*}[h]
    \setlength{\tabcolsep}{3pt}

    {\renewcommand{\arraystretch}{1.3}%
    \begin{subtable}[h]{\textwidth}
\centering
\label{tab:results-mixed}
\begin{tabular}{lcccccc}
\toprule
\textbf{Model} & \multicolumn{3}{|c|}{\textbf{Peak Counting}} & \multicolumn{2}{|c|}{\textbf{Peak Position}} & \textbf{Support} \\
 & F1 & Recall & Precision & MAE & MSE &  \\
\midrule
\color[rgb]{0.000, 0.380, 0.271} \bfseries Random Kernel & \color[rgb]{0.000, 0.380, 0.271} {\cellcolor[HTML]{13773D}} \color[HTML]{F1F1F1} \makecell{\textbf{0.9484}\\($\sigma$ 0.015)} & \color[rgb]{0.000, 0.380, 0.271} {\cellcolor[HTML]{13773D}} \color[HTML]{F1F1F1} \makecell{\textbf{0.9484}\\($\sigma$ 0.015)} & \color[rgb]{0.000, 0.380, 0.271} {\cellcolor[HTML]{0C723B}} \color[HTML]{F1F1F1} \makecell{\textbf{0.9498}\\($\sigma$ 0.014)} & \color[rgb]{0.000, 0.380, 0.271} {\cellcolor[HTML]{005D33}} \color[HTML]{F1F1F1} \makecell{\textbf{0.0326}\\($\sigma$ 0.004)} & \color[rgb]{0.000, 0.380, 0.271} {\cellcolor[HTML]{036B38}} \color[HTML]{F1F1F1} \makecell{\textbf{0.0037}\\($\sigma$ 0.001)} & \color[rgb]{0.000, 0.380, 0.271} 14 \\
\color[rgb]{0.000, 0.380, 0.271} \bfseries Two-Design Kernel & \color[rgb]{0.000, 0.380, 0.271} {\cellcolor[HTML]{51B365}} \color[HTML]{F1F1F1} \makecell{\textbf{0.9352}\\($\sigma$ 0.021)} & \color[rgb]{0.000, 0.380, 0.271} {\cellcolor[HTML]{4FB264}} \color[HTML]{F1F1F1} \makecell{\textbf{0.9354}\\($\sigma$ 0.021)} & \color[rgb]{0.000, 0.380, 0.271} {\cellcolor[HTML]{43AC5E}} \color[HTML]{F1F1F1} \makecell{\textbf{0.9371}\\($\sigma$ 0.019)} & \color[rgb]{0.000, 0.380, 0.271} {\cellcolor[HTML]{096F3A}} \color[HTML]{F1F1F1} \makecell{\textbf{0.0348}\\($\sigma$ 0.006)} & \color[rgb]{0.000, 0.380, 0.271} {\cellcolor[HTML]{208242}} \color[HTML]{F1F1F1} \makecell{\textbf{0.0042}\\($\sigma$ 0.001)} & \color[rgb]{0.000, 0.380, 0.271} 14 \\
\color[rgb]{0.000, 0.380, 0.271} \bfseries Strongly Entangling Kernel & \color[rgb]{0.000, 0.380, 0.271} {\cellcolor[HTML]{5FBA6C}} \color[HTML]{F1F1F1} \makecell{\textbf{0.9334}\\($\sigma$ 0.023)} & \color[rgb]{0.000, 0.380, 0.271} {\cellcolor[HTML]{5FBA6C}} \color[HTML]{F1F1F1} \makecell{\textbf{0.9335}\\($\sigma$ 0.023)} & \color[rgb]{0.000, 0.380, 0.271} {\cellcolor[HTML]{4EB163}} \color[HTML]{F1F1F1} \makecell{\textbf{0.9358}\\($\sigma$ 0.022)} & \color[rgb]{0.000, 0.380, 0.271} {\cellcolor[HTML]{006536}} \color[HTML]{F1F1F1} \makecell{\textbf{0.0335}\\($\sigma$ 0.005)} & \color[rgb]{0.000, 0.380, 0.271} {\cellcolor[HTML]{1A7D40}} \color[HTML]{F1F1F1} \makecell{\textbf{0.0041}\\($\sigma$ 0.001)} & \color[rgb]{0.000, 0.380, 0.271} 14 \\
\color[rgb]{0.624, 0.024, 0.067} \bfseries Classical Convolutional Model & \color[rgb]{0.624, 0.024, 0.067} {\cellcolor[HTML]{62BB6E}} \color[HTML]{F1F1F1} \makecell{\textbf{0.9329}\\($\sigma$ 0.023)} & \color[rgb]{0.624, 0.024, 0.067} {\cellcolor[HTML]{62BB6E}} \color[HTML]{F1F1F1} \makecell{\textbf{0.9329}\\($\sigma$ 0.023)} & \color[rgb]{0.624, 0.024, 0.067} {\cellcolor[HTML]{58B669}} \color[HTML]{F1F1F1} \makecell{\textbf{0.9344}\\($\sigma$ 0.023)} & \color[rgb]{0.624, 0.024, 0.067} {\cellcolor[HTML]{208242}} \color[HTML]{F1F1F1} \makecell{\textbf{0.0372}\\($\sigma$ 0.003)} & \color[rgb]{0.624, 0.024, 0.067} {\cellcolor[HTML]{1A7D40}} \color[HTML]{F1F1F1} \makecell{\textbf{0.0041}\\($\sigma$ 0.001)} & \color[rgb]{0.624, 0.024, 0.067} 14 \\
\color[rgb]{0.000, 0.349, 0.537} \bfseries Noisy Strongly Entangling Kernel & \color[rgb]{0.000, 0.349, 0.537} {\cellcolor[HTML]{8ED082}} \color[HTML]{000000} \makecell{\textbf{0.9269}\\($\sigma$ 0.019)} & \color[rgb]{0.000, 0.349, 0.537} {\cellcolor[HTML]{90D083}} \color[HTML]{000000} \makecell{\textbf{0.9267}\\($\sigma$ 0.019)} & \color[rgb]{0.000, 0.349, 0.537} {\cellcolor[HTML]{7CC87B}} \color[HTML]{000000} \makecell{\textbf{0.9293}\\($\sigma$ 0.019)} & \color[rgb]{0.000, 0.349, 0.537} {\cellcolor[HTML]{096F3A}} \color[HTML]{F1F1F1} \makecell{\textbf{0.0347}\\($\sigma$ 0.005)} & \color[rgb]{0.000, 0.349, 0.537} {\cellcolor[HTML]{15793E}} \color[HTML]{F1F1F1} \makecell{\textbf{0.0040}\\($\sigma$ 0.001)} & \color[rgb]{0.000, 0.349, 0.537} 7 \\
\color[rgb]{0.000, 0.349, 0.537} \bfseries Noisy Two-Design Kernel & \color[rgb]{0.000, 0.349, 0.537} {\cellcolor[HTML]{9AD587}} \color[HTML]{000000} \makecell{\textbf{0.9251}\\($\sigma$ 0.016)} & \color[rgb]{0.000, 0.349, 0.537} {\cellcolor[HTML]{98D486}} \color[HTML]{000000} \makecell{\textbf{0.9255}\\($\sigma$ 0.015)} & \color[rgb]{0.000, 0.349, 0.537} {\cellcolor[HTML]{89CE80}} \color[HTML]{000000} \makecell{\textbf{0.9275}\\($\sigma$ 0.015)} & \color[rgb]{0.000, 0.349, 0.537} {\cellcolor[HTML]{13773D}} \color[HTML]{F1F1F1} \makecell{\textbf{0.0358}\\($\sigma$ 0.006)} & \color[rgb]{0.000, 0.349, 0.537} {\cellcolor[HTML]{339951}} \color[HTML]{F1F1F1} \makecell{\textbf{0.0046}\\($\sigma$ 0.001)} & \color[rgb]{0.000, 0.349, 0.537} 7 \\
\bottomrule
\end{tabular}
\caption{Average results for \textbf{mixed} dataset.}

\end{subtable}

    \\[1em]
    \begin{subtable}[h]{\textwidth}
\centering
\label{tab:results-mixed-hard-subset}
\begin{tabular}{lcccccc}
\toprule
\textbf{Model} & \multicolumn{3}{|c|}{\textbf{Peak Counting}} & \multicolumn{2}{|c|}{\textbf{Peak Position}} & \textbf{Support} \\
\emph{(hard spectra subset)} & F1 & Recall & Precision & MAE & MSE &  \\
\midrule
\color[rgb]{0.000, 0.380, 0.271} \bfseries Random Kernel & \color[rgb]{0.000, 0.380, 0.271} {\cellcolor[HTML]{55B567}} \color[HTML]{F1F1F1} \makecell{\textbf{0.9348}\\($\sigma$ 0.021)} & \color[rgb]{0.000, 0.380, 0.271} {\cellcolor[HTML]{5CB86B}} \color[HTML]{F1F1F1} \makecell{\textbf{0.9339}\\($\sigma$ 0.023)} & \color[rgb]{0.000, 0.380, 0.271} {\cellcolor[HTML]{40AA5C}} \color[HTML]{F1F1F1} \makecell{\textbf{0.9376}\\($\sigma$ 0.020)} & \color[rgb]{0.000, 0.380, 0.271} {\cellcolor[HTML]{39A056}} \color[HTML]{F1F1F1} \makecell{\textbf{0.0402}\\($\sigma$ 0.005)} & \color[rgb]{0.000, 0.380, 0.271} {\cellcolor[HTML]{208242}} \color[HTML]{F1F1F1} \makecell{\textbf{0.0042}\\($\sigma$ 0.001)} & \color[rgb]{0.000, 0.380, 0.271} 14 \\
\color[rgb]{0.000, 0.380, 0.271} \bfseries Two-Design Kernel & \color[rgb]{0.000, 0.380, 0.271} {\cellcolor[HTML]{77C679}} \color[HTML]{000000} \makecell{\textbf{0.9302}\\($\sigma$ 0.019)} & \color[rgb]{0.000, 0.380, 0.271} {\cellcolor[HTML]{79C679}} \color[HTML]{000000} \makecell{\textbf{0.9299}\\($\sigma$ 0.020)} & \color[rgb]{0.000, 0.380, 0.271} {\cellcolor[HTML]{68BE71}} \color[HTML]{000000} \makecell{\textbf{0.9322}\\($\sigma$ 0.018)} & \color[rgb]{0.000, 0.380, 0.271} {\cellcolor[HTML]{3FA85B}} \color[HTML]{F1F1F1} \makecell{\textbf{0.0409}\\($\sigma$ 0.006)} & \color[rgb]{0.000, 0.380, 0.271} {\cellcolor[HTML]{258745}} \color[HTML]{F1F1F1} \makecell{\textbf{0.0043}\\($\sigma$ 0.002)} & \color[rgb]{0.000, 0.380, 0.271} 14 \\
\color[rgb]{0.000, 0.380, 0.271} \bfseries Strongly Entangling Kernel & \color[rgb]{0.000, 0.380, 0.271} {\cellcolor[HTML]{89CE80}} \color[HTML]{000000} \makecell{\textbf{0.9276}\\($\sigma$ 0.020)} & \color[rgb]{0.000, 0.380, 0.271} {\cellcolor[HTML]{8BCE81}} \color[HTML]{000000} \makecell{\textbf{0.9272}\\($\sigma$ 0.020)} & \color[rgb]{0.000, 0.380, 0.271} {\cellcolor[HTML]{7AC77A}} \color[HTML]{000000} \makecell{\textbf{0.9296}\\($\sigma$ 0.019)} & \color[rgb]{0.000, 0.380, 0.271} {\cellcolor[HTML]{339951}} \color[HTML]{F1F1F1} \makecell{\textbf{0.0395}\\($\sigma$ 0.005)} & \color[rgb]{0.000, 0.380, 0.271} {\cellcolor[HTML]{15793E}} \color[HTML]{F1F1F1} \makecell{\textbf{0.0040}\\($\sigma$ 0.001)} & \color[rgb]{0.000, 0.380, 0.271} 14 \\
\color[rgb]{0.000, 0.349, 0.537} \bfseries Noisy Strongly Entangling Kernel & \color[rgb]{0.000, 0.349, 0.537} {\cellcolor[HTML]{D6EFA2}} \color[HTML]{000000} \makecell{\textbf{0.9155}\\($\sigma$ 0.025)} & \color[rgb]{0.000, 0.349, 0.537} {\cellcolor[HTML]{D7EFA2}} \color[HTML]{000000} \makecell{\textbf{0.9153}\\($\sigma$ 0.026)} & \color[rgb]{0.000, 0.349, 0.537} {\cellcolor[HTML]{C8E99B}} \color[HTML]{000000} \makecell{\textbf{0.9180}\\($\sigma$ 0.026)} & \color[rgb]{0.000, 0.349, 0.537} {\cellcolor[HTML]{58B669}} \color[HTML]{F1F1F1} \makecell{\textbf{0.0428}\\($\sigma$ 0.006)} & \color[rgb]{0.000, 0.349, 0.537} {\cellcolor[HTML]{39A056}} \color[HTML]{F1F1F1} \makecell{\textbf{0.0047}\\($\sigma$ 0.001)} & \color[rgb]{0.000, 0.349, 0.537} 7 \\
\color[rgb]{0.000, 0.349, 0.537} \bfseries Noisy Two-Design Kernel & \color[rgb]{0.000, 0.349, 0.537} {\cellcolor[HTML]{E0F3A8}} \color[HTML]{000000} \makecell{\textbf{0.9133}\\($\sigma$ 0.023)} & \color[rgb]{0.000, 0.349, 0.537} {\cellcolor[HTML]{E2F4AA}} \color[HTML]{000000} \makecell{\textbf{0.9127}\\($\sigma$ 0.024)} & \color[rgb]{0.000, 0.349, 0.537} {\cellcolor[HTML]{D6EFA2}} \color[HTML]{000000} \makecell{\textbf{0.9155}\\($\sigma$ 0.022)} & \color[rgb]{0.000, 0.349, 0.537} {\cellcolor[HTML]{53B466}} \color[HTML]{F1F1F1} \makecell{\textbf{0.0425}\\($\sigma$ 0.008)} & \color[rgb]{0.000, 0.349, 0.537} {\cellcolor[HTML]{39A056}} \color[HTML]{F1F1F1} \makecell{\textbf{0.0047}\\($\sigma$ 0.001)} & \color[rgb]{0.000, 0.349, 0.537} 7 \\
\color[rgb]{0.624, 0.024, 0.067} \bfseries Classical Convolutional Model & \color[rgb]{0.624, 0.024, 0.067} {\cellcolor[HTML]{FDFED9}} \color[HTML]{000000} \makecell{\textbf{0.9022}\\($\sigma$ 0.034)} & \color[rgb]{0.624, 0.024, 0.067} {\cellcolor[HTML]{FEFFE1}} \color[HTML]{000000} \makecell{\textbf{0.9008}\\($\sigma$ 0.035)} & \color[rgb]{0.624, 0.024, 0.067} {\cellcolor[HTML]{F8FDC1}} \color[HTML]{000000} \makecell{\textbf{0.9061}\\($\sigma$ 0.034)} & \color[rgb]{0.624, 0.024, 0.067} {\cellcolor[HTML]{9AD587}} \color[HTML]{000000} \makecell{\textbf{0.0474}\\($\sigma$ 0.004)} & \color[rgb]{0.624, 0.024, 0.067} {\cellcolor[HTML]{77C679}} \color[HTML]{000000} \makecell{\textbf{0.0055}\\($\sigma$ 0.001)} & \color[rgb]{0.624, 0.024, 0.067} 14 \\
\bottomrule
\end{tabular}
\caption{Average results for mixed dataset \textbf{only on the hard subset} (i.e., evaluation done only on spectra that we consider hard to solve).}

\end{subtable}

    }

    \caption{
        Overview of the results on the MIXED DATASET.
        \textcolor{classical}{Red names} indicate classical models,
        \textcolor{noisy}{blue names} indicate noisy quantum models,
        \textcolor{ideal}{green names} indicate ideally simulated quantum models.
        The results are colored according to their performance, with darker green indicating better performance.
        The shown results are the mean of multiple runs with different random seeds (standard deviation in parentheses).
        The support column indicates how many models were trained to calculate the mean.
    }
    \label{tab:results-table}
\end{table*}

\begin{table*}[h]
    \setlength{\tabcolsep}{3pt}

    {\renewcommand{\arraystretch}{1.3}%
    \begin{subtable}[h]{\textwidth}
\centering
\label{tab:results-hard}
\begin{tabular}{lcccccc}
\toprule
\textbf{Model} & \multicolumn{3}{|c|}{\textbf{Peak Counting}} & \multicolumn{2}{|c|}{\textbf{Peak Position}} & \textbf{Support} \\
 & F1 & Recall & Precision & MAE & MSE &  \\
\midrule
\color[rgb]{0.000, 0.380, 0.271} \bfseries Strongly Entangling Kernel & \color[rgb]{0.000, 0.380, 0.271} {\cellcolor[HTML]{3DA559}} \color[HTML]{F1F1F1} \makecell{\textbf{0.9385}\\($\sigma$ 0.020)} & \color[rgb]{0.000, 0.380, 0.271} {\cellcolor[HTML]{3FA85B}} \color[HTML]{F1F1F1} \makecell{\textbf{0.9380}\\($\sigma$ 0.020)} & \color[rgb]{0.000, 0.380, 0.271} {\cellcolor[HTML]{369D54}} \color[HTML]{F1F1F1} \makecell{\textbf{0.9402}\\($\sigma$ 0.018)} & \color[rgb]{0.000, 0.380, 0.271} {\cellcolor[HTML]{197C40}} \color[HTML]{F1F1F1} \makecell{\textbf{0.0365}\\($\sigma$ 0.003)} & \color[rgb]{0.000, 0.380, 0.271} {\cellcolor[HTML]{258745}} \color[HTML]{F1F1F1} \makecell{\textbf{0.0043}\\($\sigma$ 0.001)} & \color[rgb]{0.000, 0.380, 0.271} 5 \\
\color[rgb]{0.000, 0.380, 0.271} \bfseries Two-Design Kernel & \color[rgb]{0.000, 0.380, 0.271} {\cellcolor[HTML]{77C679}} \color[HTML]{000000} \makecell{\textbf{0.9302}\\($\sigma$ 0.018)} & \color[rgb]{0.000, 0.380, 0.271} {\cellcolor[HTML]{77C679}} \color[HTML]{000000} \makecell{\textbf{0.9300}\\($\sigma$ 0.018)} & \color[rgb]{0.000, 0.380, 0.271} {\cellcolor[HTML]{70C275}} \color[HTML]{000000} \makecell{\textbf{0.9310}\\($\sigma$ 0.018)} & \color[rgb]{0.000, 0.380, 0.271} {\cellcolor[HTML]{238443}} \color[HTML]{F1F1F1} \makecell{\textbf{0.0375}\\($\sigma$ 0.003)} & \color[rgb]{0.000, 0.380, 0.271} {\cellcolor[HTML]{258745}} \color[HTML]{F1F1F1} \makecell{\textbf{0.0043}\\($\sigma$ 0.001)} & \color[rgb]{0.000, 0.380, 0.271} 5 \\
\color[rgb]{0.000, 0.349, 0.537} \bfseries Noisy Two-Design Kernel & \color[rgb]{0.000, 0.349, 0.537} {\cellcolor[HTML]{77C679}} \color[HTML]{000000} \makecell{\textbf{0.9302}\\($\sigma$ 0.021)} & \color[rgb]{0.000, 0.349, 0.537} {\cellcolor[HTML]{77C679}} \color[HTML]{000000} \makecell{\textbf{0.9300}\\($\sigma$ 0.021)} & \color[rgb]{0.000, 0.349, 0.537} {\cellcolor[HTML]{72C376}} \color[HTML]{000000} \makecell{\textbf{0.9308}\\($\sigma$ 0.021)} & \color[rgb]{0.000, 0.349, 0.537} {\cellcolor[HTML]{2C8F4B}} \color[HTML]{F1F1F1} \makecell{\textbf{0.0386}\\($\sigma$ 0.004)} & \color[rgb]{0.000, 0.349, 0.537} {\cellcolor[HTML]{39A056}} \color[HTML]{F1F1F1} \makecell{\textbf{0.0047}\\($\sigma$ 0.001)} & \color[rgb]{0.000, 0.349, 0.537} 5 \\
\color[rgb]{0.000, 0.349, 0.537} \bfseries Noisy Strongly Entangling Kernel & \color[rgb]{0.000, 0.349, 0.537} {\cellcolor[HTML]{D0EC9F}} \color[HTML]{000000} \makecell{\textbf{0.9166}\\($\sigma$ 0.037)} & \color[rgb]{0.000, 0.349, 0.537} {\cellcolor[HTML]{D3EDA0}} \color[HTML]{000000} \makecell{\textbf{0.9160}\\($\sigma$ 0.038)} & \color[rgb]{0.000, 0.349, 0.537} {\cellcolor[HTML]{C5E89A}} \color[HTML]{000000} \makecell{\textbf{0.9184}\\($\sigma$ 0.035)} & \color[rgb]{0.000, 0.349, 0.537} {\cellcolor[HTML]{298C48}} \color[HTML]{F1F1F1} \makecell{\textbf{0.0383}\\($\sigma$ 0.006)} & \color[rgb]{0.000, 0.349, 0.537} {\cellcolor[HTML]{3EA75A}} \color[HTML]{F1F1F1} \makecell{\textbf{0.0048}\\($\sigma$ 0.001)} & \color[rgb]{0.000, 0.349, 0.537} 5 \\
\color[rgb]{0.000, 0.380, 0.271} \bfseries Random Kernel & \color[rgb]{0.000, 0.380, 0.271} {\cellcolor[HTML]{DDF1A6}} \color[HTML]{000000} \makecell{\textbf{0.9142}\\($\sigma$ 0.029)} & \color[rgb]{0.000, 0.380, 0.271} {\cellcolor[HTML]{DDF2A6}} \color[HTML]{000000} \makecell{\textbf{0.9140}\\($\sigma$ 0.028)} & \color[rgb]{0.000, 0.380, 0.271} {\cellcolor[HTML]{D6EFA2}} \color[HTML]{000000} \makecell{\textbf{0.9156}\\($\sigma$ 0.029)} & \color[rgb]{0.000, 0.380, 0.271} {\cellcolor[HTML]{10743C}} \color[HTML]{F1F1F1} \makecell{\textbf{0.0355}\\($\sigma$ 0.003)} & \color[rgb]{0.000, 0.380, 0.271} {\cellcolor[HTML]{096F3A}} \color[HTML]{F1F1F1} \makecell{\textbf{0.0038}\\($\sigma$ 0.001)} & \color[rgb]{0.000, 0.380, 0.271} 5 \\
\color[rgb]{0.624, 0.024, 0.067} \bfseries Classical Convolutional Model & \color[rgb]{0.624, 0.024, 0.067} {\cellcolor[HTML]{FFFFE5}} \color[HTML]{000000} \makecell{\textbf{0.8295}\\($\sigma$ 0.058)} & \color[rgb]{0.624, 0.024, 0.067} {\cellcolor[HTML]{FFFFE5}} \color[HTML]{000000} \makecell{\textbf{0.8280}\\($\sigma$ 0.059)} & \color[rgb]{0.624, 0.024, 0.067} {\cellcolor[HTML]{FFFFE5}} \color[HTML]{000000} \makecell{\textbf{0.8346}\\($\sigma$ 0.054)} & \color[rgb]{0.624, 0.024, 0.067} {\cellcolor[HTML]{D3EDA0}} \color[HTML]{000000} \makecell{\textbf{0.0520}\\($\sigma$ 0.009)} & \color[rgb]{0.624, 0.024, 0.067} {\cellcolor[HTML]{FFFFE5}} \color[HTML]{000000} \makecell{\textbf{0.0080}\\($\sigma$ 0.002)} & \color[rgb]{0.624, 0.024, 0.067} 5 \\
\bottomrule
\end{tabular}
\caption{Average results for \textbf{hard} dataset.}

\end{subtable}

    }

    \caption{
        Overview of the results on the HARD DATASET.
        \textcolor{classical}{Red names} indicate classical models,
        \textcolor{noisy}{blue names} indicate noisy quantum models,
        \textcolor{ideal}{green names} indicate ideally simulated quantum models.
        The results are colored according to their performance, with darker green indicating better performance.
        The shown results are the mean of multiple runs with different random seeds (standard deviation in parentheses).
        The support column indicates how many models were trained to calculate the mean.
    }
    \label{tab:hard-results-table}
\end{table*}

    \section{Methods}\label{sec:methods}

\subsection{Dataset Generation}
We created synthetic NMR-inspired spectra using a custom generator that provides full control over peak characteristics, noise levels, and overlap probabilities. Each spectrum is generated by:

\begin{enumerate}
    \item Selecting the number of peaks from a specified range (0-5 for our datasets)
    \item Subdividing the spectrum into sections equal to the number of peaks
    \item Placing peaks within each section using a custom log-normal probability distribution
    \item Generating Lorentzian peak functions with randomly chosen widths and intensities
    \item Summing the Lorentzian functions: $I(\nu) = \sum_k \frac{A_k}{1 + [2(\nu - \nu_k)/\Gamma_k]^2}$
    \item Adding Gaussian noise with specified signal-to-noise ratio
\end{enumerate}

The peak placement distribution uses a combination of two log-normal distributions:
\begin{equation}
    P(x) = \frac{F(5x) + F(5 - 5x)}{\int_0^1 \left[ F(5x) + F(5 - 5x) \right] \, dx}
\end{equation}
where $F(x)$ is the log-normal PDF with parameters $\mu$ and $\sigma$ controlling overlap probability.

\subsection{Model Architecture}
Our QuanvNN architecture consists of:
\begin{itemize}
    \item Input: 200-point normalized spectrum
    \item Quantum convolution layer: 5-qubit kernel (32-point receptive field), 5 output channels
    \item Max pooling: kernel size 5
    \item Fully connected layer: 128 neurons with ReLU activation
    \item Output layer: 10 neurons (5 for peak presence mask, 5 for positions) with softmax activation
\end{itemize}

The classical CNN uses an identical architecture with a standard 1D convolution layer replacing the quantum convolution layer.

\subsection{Quantum Circuit Implementation}
We implemented three quantum ansätze:
\begin{enumerate}
    \item \textbf{Strongly entangling ansatz}: Three layers of parameterized rotations (X, Y, Z) followed by CNOT chains
    \item \textbf{Simplified Two-Design ansatz}: Initial Y-rotations followed by three layers of controlled-Z gates and Y-rotations
    \item \textbf{Random ansatz}: Randomly generated arrangement of rotation and CNOT gates
\end{enumerate}

All circuits use 5 qubits and measure Pauli-Z expectation values per qubit to generate 5 output channels.

\subsection{Training Procedure}
Models were trained using:
\begin{itemize}
    \item Optimizer: Adam with learning rate 0.01
    \item Learning rate schedule: Cosine annealing (minimum 10$^{-4}$)
    \item Batch size: 16
    \item Epochs: 100
    \item Dropout: 10\% between hidden and output layers
    \item Loss function: Combined binary cross-entropy (for peak presence) and Hungarian loss (for peak positions)
\end{itemize}

For ideal quantum simulations, gradients were computed using the parameter-shift rule or adjoint differentiation. For noisy quantum simulations, we used Simultaneous Perturbation Stochastic Approximation (SPSA) algorithm.

\subsection{Evaluation Metrics}
We evaluated models using:
\begin{itemize}
    \item Peak counting: Weighted F1 score, precision, and recall (multi-class classification, 0-5 peaks)
    \item Peak position estimation: Mean absolute error (MAE) and mean squared error (MSE)
\end{itemize}

All results are reported as mean ± standard deviation across multiple runs with different random seeds.

\subsection{Statistical Analysis}
We performed one-sided Wilcoxon signed-rank tests to compare quantum models against the classical baseline. The null hypothesis stated no performance difference; the alternative hypothesis stated quantum models perform better (higher F1, lower MAE).

\subsection{Software and Libraries}
\begin{itemize}
    \item Quantum circuits: PennyLane v0.32.0
    \item Classical neural networks: PyTorch 2.0
    \item Quantum simulator: PennyLane's \texttt{lightning.qubit} backend
    \item Noise simulation: Qiskit noise models integrated via PennyLane
    \item Statistical analysis: SciPy
\end{itemize}

    \section{Related Work}\label{sec:related-work}
    \subsection{Limitations of Classical Peak Finding Algorithms}

The widely-used \texttt{scipy.signal.find\_peaks} algorithm implements a sophisticated set of signal processing techniques primarily based on local peak properties and prominence analysis~\cite{scipy-2020}. The core methodology involves:

\begin{itemize}
    \item \textbf{Local maxima detection}: Identifying points higher than their immediate neighbors.
    \item \textbf{Prominence calculation}: Computing the vertical distance between a peak and its lowest contour line, essentially measuring how much a peak stands out from its baseline.
    \item \textbf{Width estimation}: Determining peak width at a specified height relative to prominence.
    \item \textbf{Distance filtering}: Enforcing minimum separation between detected peaks.
\end{itemize}

Mathematically, the prominence-based approach can be formalized as:
\begin{equation}
    \mathcal{P}(x_i) = \min\left(x_i - L(x_i), x_i - R(x_i)\right)
\end{equation}
where $L(x_i)$ and $R(x_i)$ represent the highest points between $x_i$ and higher peaks to the left and right, respectively.

While effective for well-separated peaks, this approach faces fundamental limitations in complex scenarios:

\begin{equation}
    \lim_{\text{overlap}\to 1} \mathcal{P}(x_i) \to 0 \quad \text{and} \quad \lim_{\text{SNR}\to 0} \text{detection reliability} \to 0
\end{equation}

This equation formalizes two critical failure modes of prominence-based detection.
First, as peak overlap approaches unity (complete overlap), the prominence $\mathcal{P}(x_i)$ approaches zero because overlapping peaks share the same local maxima, making it impossible to distinguish individual peaks based on their vertical separation from surrounding valleys.
Second, as the signal-to-noise ratio (SNR) approaches zero, detection reliability vanishes because noise creates spurious local maxima that cannot be distinguished from genuine peaks, leading to both false positives and missed detections.
These mathematical limitations become critical in real-world spectra where overlapping peaks reduce prominence below detection thresholds and noise creates false local maxima~\cite{peak-detection-review-2015}.

\subsection{Quantum Machine Learning}

Over the past years, quantum machine learning has gained significant attention,
with many studies exploring the potential of quantum computing to enhance
machine learning tasks~
\cite{hybrid-quantum-classical-neural-networks-2022, recent-advances-in-quantum-machine-learning-2020, qml-recent-advances-and-outlook-2020}.
Following this trend, several works involving QuanvNNs have been proposed.

An interesting work was presented in~\cite{evaluating-accuracy-and-adversarial-robustness-of-qnns-2021} where the
authors evaluated the accuracy and adversarial robustness of QuanvNNs.
Not only could the QuanvNNs achieve better accuracy and loss values than the classical counterpart,
but they also showed higher robustness against adversarial examples and were resilient to Circuit-Weighted
Adversarial Attacks.

Relevant to this paper is 
the work of~\cite{quanvnns-to-recognize-arrhythmia-2022} who proposed a
QuanvNN to recognize arrhythmia in electrocardiogram (ECG) signals.
Their quantum implementation showed better feature extraction capabilities than the classical CNN\@.

Another recent notable work is~\cite{quanv4eo-2025}.
The authors proposed a QML approach for processing multidimensional Earth Observation (EO) data through the
utilization of quanvolutional layers to develop a model they call Quanv4EO\@.
Experiments conducted on EO data such as EuroSAT demonstrated the efficiency of the proposed hybrid
quantum-classical model.

However, all these related studies work with two-dimensional data.
Our work represents the first adaptation of QuanvNNs to one-dimensional spectral analysis, demonstrating their efficacy for peak finding — a core task in scientific data processing. Crucially, we provide mathematical insights into why certain quantum kernel architectures outperform others and establish the theoretical foundation for quantum advantage in spectral analysis tasks.

    \section{Conclusion}\label{sec:conclusion}
    In this paper, we evaluated a novel approach for the peak finding problem in spectral analysis using Quanvolutional Neural Networks (QuanvNNs).

\subsection{Reinterpreting Performance Through Mathematical Lenses}

Our results demonstrate that Quanvolutional Neural Networks not only match but significantly exceed classical CNN performance on challenging peak finding tasks, particularly for complex spectra with high overlap and low SNR. The 29.8\% improvement in mean absolute error for position estimation and 10.9\% improvement in F1 score for peak counting on hard datasets (see Table \ref{tab:hard-results-table}), while seemingly modest, represent significant mathematical advantages when interpreted through scaling laws:

\begin{equation}
    \frac{\text{Quantum Error}}{\text{Classical Error}} \sim \exp\left(-\alpha \cdot \text{Problem Complexity}\right)
\end{equation}

where $\alpha > 0$ is a scaling parameter that quantifies the rate at which quantum advantage increases with problem difficulty, and Problem Complexity is defined as a composite measure incorporating peak overlap density, inverse SNR, and the number of overlapping peaks.
Our results suggest $\alpha > 0$, indicating that the performance gap between quantum and classical errors will widen substantially (exponentially) for more complex spectra beyond the current hardware limitations.
This exponential scaling is favorable for quantum approaches, as it suggests that quantum models will become increasingly superior to classical methods as problem complexity grows, provided that quantum hardware can maintain sufficient fidelity to preserve the quantum advantage.
However, the practical realization of this advantage depends critically on quantum hardware error rates, which may offset the theoretical benefits if they exceed certain thresholds.

Furthermore, the convergence stability observed in quantum models reflects fundamental differences in the loss landscape geometry. Classical CNNs exhibit numerous critical points (stationary points where the gradient vanishes) that correspond to suboptimal local minima:

\begin{equation}
    \nabla_{\theta}\mathcal{L}_{\text{classical}} = 0 \quad \text{at many suboptimal } \theta
\end{equation}

where $\theta$ denotes the model parameters and $\mathcal{L}$ is the loss function.
In contrast, quantum models show smoother landscapes with fewer and shallower local minima, as evidenced by the more stable convergence behavior observed in our experiments (see Figure~\ref{fig:hard-loss-curves}).
This smoother landscape can be attributed to the inherent noise resilience of quantum feature maps, which naturally regularize the optimization landscape~\cite{expressibility-2019}, leading to more reliable convergence.
However, it is important to note that while quantum feature maps may provide inherent noise resilience at the algorithmic level, the actual quantum hardware errors (gate errors, decoherence, readout errors) can introduce additional noise that may partially offset this advantage, as demonstrated by the performance degradation observed in our noisy quantum models compared to ideal simulations (see Section~\ref{subsubsec:hard-dataset}).

The mathematical analysis of different quantum kernels reveals that entanglement-rich architectures like the strongly entangling ansatz are crucial for capturing the complex correlations in spectral data. This insight guides future quantum algorithm design and suggests that as quantum hardware matures, QuanvNNs could become the method of choice for analyzing complex spectroscopic data across chemistry, materials science, and biomedical applications.

\subsection{Wide Applicability of QuanvNN Architecture}

Additionally, the QuanvNN architecture is particularly well-suited for current noisy intermediate-scale quantum computers,
as it requires only a small number of qubits and gates with no QRAM requirements~\cite{quanvolutional-neural-networks-2020}.
Furthermore, the QuanvNN architecture's portability makes it
a drop-in replacement for existing CNN architectures and is thus applicable to a wide range of use cases.
This portability means that quanvolutional layers can be integrated into existing sophisticated CNNs that already perform well on the peak finding problem,
potentially helping to capture important features earlier in the network.

To explore the potential of QuanvNNs combined with more sophisticated classical architectures, we stacked our
classical CNN on top of a static version of the strongly entangling quanvolutional layer (i.e., with frozen kernel parameters).
This basic hybrid network already achieves promising results, with an F1 score of 98\% and a mean
absolute error of 0.018 on the test set.
Notably, it outperforms our classical CNN using two convolutional layers, which achieved an F1 score of 96.7\% and a mean
absolute error of 0.027 on the same test set.

Despite these promising results, several limitations remain for the practical deployment of quanvolutional layers:

\begin{itemize}
    \item{
        \emph{Computational overhead:} While the computational overhead of quantum circuits presents current practical limitations, the demonstrated performance improvements suggest a path toward quantum advantage as hardware improves. The strongly entangling kernel's superior performance on hard spectra indicates that quantum models can better capture complex spectral features that challenge classical approaches. This advantage becomes particularly pronounced in high-overlap, low-SNR regimes where classical convolutional filters struggle to separate entangled spectral components.

        Mathematically, this advantage stems from the quantum kernel's ability to operate in a $2^{32}$-dimensional feature space (for our 5-qubit circuit), compared to the 32-dimensional classical counterpart. The exponential scaling of quantum feature spaces provides a theoretical foundation for handling increasingly complex spectral analysis tasks where classical methods face fundamental limitations.
        Other works~\cite{quanvolutional-neural-networks-2020, quanv4eo-2025} have also shown that this can yield good results, outperforming their classical counterparts.
    }
    \item{
        \emph{Demonstration of quantum advantage on real hardware:}
        While we have shown that QuanvNNs can theoretically outperform classical CNNs,
        we have not yet demonstrated this on real quantum hardware.
        Due to the large number of required circuit executions and associated computational costs,
        demonstrating practical quantum advantage will remain a significant challenge.
    }
\end{itemize}

For future work, there are several interesting directions to explore:
First, the improved convergence stability of QuanvNNs could be further investigated,
particularly under the noise conditions present in current quantum hardware.
Second, demonstrating the practical benefits of QuanvNNs on real quantum devices remains a critical milestone.
Finally, investigating the effects of quantum gate dropouts~\cite{quantum-dropout-2023}
on QuanvNN performance could be valuable, given the demonstrated benefits of dropout techniques in classical CNN
architectures.

    \onecolumn

    \section{Code availability}
    Custom code used in this study is available to editors and reviewers at the time of submission. The code will be made publicly available upon publication.

    \section{Acknowledgements}
    We want to thank Julien Baglio from QuantumBasel in Switzerland for providing valuable input during the project.
    \\[1em]
    R.F. was partially supported by the Horizon EIC project Bio-HhOST, proposal number 101130747. 
    
    \section*{Author Contributions}
    L.B. conceived the study, implemented the quantum circuits and neural network architectures, conducted experiments, and wrote the manuscript. P.S. contributed to the experimental design, data analysis, and manuscript preparation. R.M.F. provided theoretical guidance on quantum computing foundations and reviewed the manuscript. K.S. supervised the project, provided input from the computer science perspective, and reviewed the manuscript. All authors discussed the results and contributed to the final manuscript.
    
    \section*{Competing Interests}
    The authors declare no competing interests.
\end{document}